\documentclass[runningheads]{llncs}

\usepackage{eccv}

\usepackage{eccvabbrv}

\usepackage{graphicx}
\graphicspath{{./figures/}}
\usepackage{booktabs}
\usepackage{tcolorbox}
\usepackage{listings}
\usepackage{xcolor}
\usepackage{multirow}
\usepackage{enumitem}

\usepackage[accsupp]{axessibility}  

\usepackage{amssymb}

\usepackage{hyperref}

\usepackage{orcidlink}

\usepackage{microtype}

\AtEndPreamble{%
  \crefname{figure}{Fig.}{Figs.}%
  \Crefname{figure}{Figure}{Figures}%
}

\colorlet{punct}{red!60!black}
\definecolor{background}{HTML}{EEEEEE}
\definecolor{delim}{RGB}{20,105,176}
\colorlet{numb}{magenta!60!black}
\definecolor{codegreen}{rgb}{0,0.6,0}
\definecolor{codegray}{rgb}{0.5,0.5,0.5}
\definecolor{codepurple}{rgb}{0.58,0,0.82}
\definecolor{backcolour}{rgb}{0.95,0.95,0.92}

\lstdefinelanguage{json}{
    basicstyle=\normalfont\ttfamily,
    numbers=left,
    numberstyle=\scriptsize,
    stepnumber=1,
    numbersep=8pt,
    showstringspaces=false,
    breaklines=true,
    frame=lines,
    backgroundcolor=\color{background},
    literate=
     *{0}{{{\color{numb}0}}}{1}
      {1}{{{\color{numb}1}}}{1}
      {2}{{{\color{numb}2}}}{1}
      {3}{{{\color{numb}3}}}{1}
      {4}{{{\color{numb}4}}}{1}
      {5}{{{\color{numb}5}}}{1}
      {6}{{{\color{numb}6}}}{1}
      {7}{{{\color{numb}7}}}{1}
      {8}{{{\color{numb}8}}}{1}
      {9}{{{\color{numb}9}}}{1}
      {:}{{{\color{punct}{:}}}}{1}
      {,}{{{\color{punct}{,}}}}{1}
      {\{}{{{\color{punct}{\{}}}}{1}
      {\}}{{{\color{punct}{\}}}}}{1}
      {[}{{{\color{punct}{[}}}}{1}
      {]}{{{\color{punct}{]}}}}{1},
}

\lstdefinestyle{mystyle}{
    backgroundcolor=\color{backcolour},
    commentstyle=\color{codegreen},
    keywordstyle=\color{magenta},
    numberstyle=\tiny\color{codegray},
    stringstyle=\color{codepurple},
    basicstyle=\ttfamily\footnotesize,
    breakatwhitespace=false,
    breaklines=true,
    captionpos=b,
    keepspaces=true,
    numbers=left,
    numbersep=5pt,
    showspaces=false,
    showstringspaces=false,
    showtabs=false,
    tabsize=2
}
\begin{document}

\title{GH-ESD: Grounded Hypothesis-Driven Error Slice Discovery for Instance-Level Vision Tasks}

\titlerunning{GH-ESD: Grounded Hypothesis-Driven Error Slice Discovery}

\author{Wei Zhang\inst{1}\thanks{These authors contributed equally to this research.} \and
Chaoqun Wang\inst{1}$^*$ \and
Zixuan Guan\inst{1} \and
Ping Sheng Kao\inst{2} \and
Pengfei Zhao\inst{1} \and
Peng Wu\inst{1} \and
Sifeng He\inst{1}\thanks{Corresponding Author, Project Lead}}

\authorrunning{W. Zhang et al.}

\institute{
Apple, Beijing, China \\
\email{\{wzhang52, chaoqun\_wang, zixuan\_guan, pzhao23, pwu4, he\_sifeng\}@apple.com}
\and
Apple, Cupertino, CA, USA \\
\email{pingsheng\_kao@apple.com}
}
\maketitle

\begin{abstract}
Systematic failures of vision models on semantically coherent subsets, known as error slices, reveal limitations in robustness and evaluation.
Existing slice discovery approaches largely model slices as clusters in representation space or combinations of predefined attributes. While effective for image-level classification, such formulations are insufficient for instance-level tasks such as object detection and segmentation, where failures often arise from contextual, relational, and spatially grounded visual patterns.
We propose GH-ESD (Grounded Hypothesis-Driven Error Slice Discovery), a generate-and-verify framework that reformulates slice discovery as grounded hypothesis generation and statistical verification. GH-ESD constructs relational failure hypotheses using LLM priors and grounded visual evidence, discovers hypothesis slices at the instance level via Vision-Language Models, and verifies them through statistical trend analysis over instance-level errors.
We also introduce GESD (Grounded Error Slice Dataset), a new benchmark for instance-level error slice discovery, providing expert-defined and spatially grounded slices derived from detection and segmentation failures.
Extensive experiments demonstrate that GH-ESD consistently outperforms baselines, improving Precision@10 by 0.10 (0.73 vs. 0.63) on the GESD benchmark for detection tasks, while also supporting segmentation scenarios. GH-ESD identifies interpretable slices that facilitate actionable model improvements. GESD is publicly available at  \href{https://github.com/apple/ml-fine-grained-error-slice-discovery-dataset}{https://github.com/apple/ml-fine-grained-error-slice-discovery-dataset}.

\end{abstract}
    
\section{Introduction}
\label{sec:intro}
Despite the remarkable performance of modern deep vision models, their reliability remains challenged by systematic failures on semantically coherent subsets of data, commonly referred to as \textit{error slices}~\cite{zhao2025hibug2}. 
These slices expose structured weaknesses in model behavior and provide critical signals for robustness analysis and model refinement, particularly in high-stakes domains such as autonomous driving~\cite{blum2021fishyscapes} and robotics~\cite{zhou2025code}. Identifying such slices is essential for trustworthy model evaluation.

\begin{figure}[tb]
  \centering
  \includegraphics[width=\linewidth]{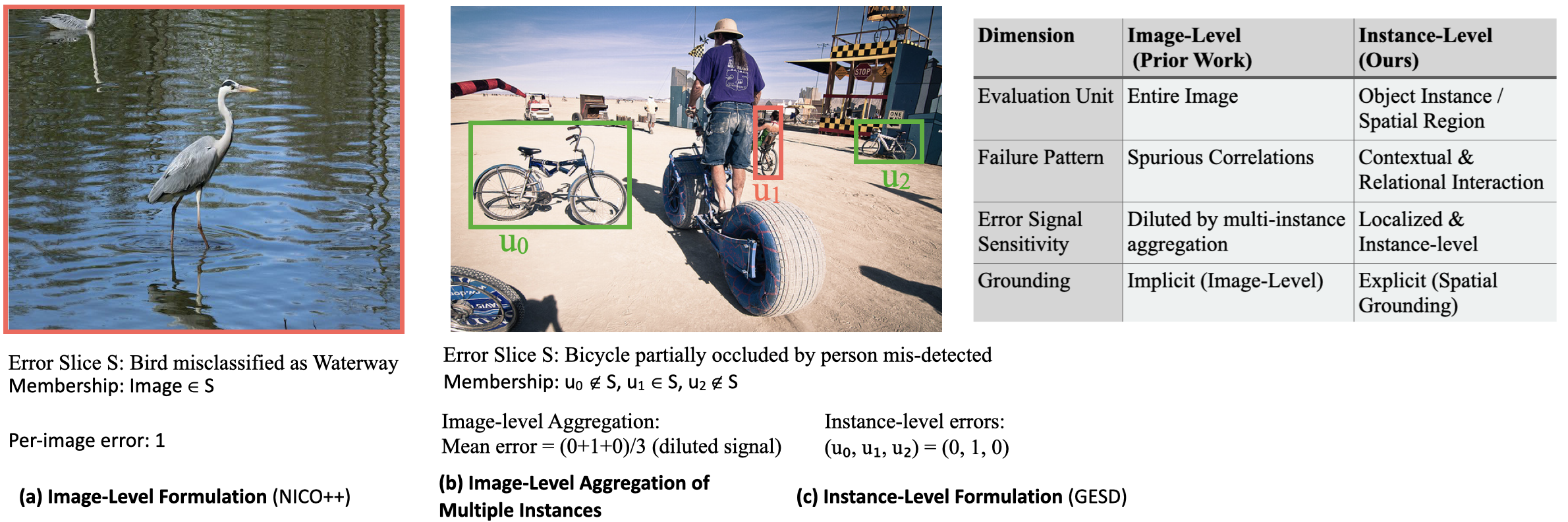}
  \caption{Comparison of image-level and instance-level error slice discovery.}
  \label{fig:introduction}
\end{figure}

Most existing slice discovery methods implicitly model slices as regions in representation space or combinations of predefined attributes~\cite{ghosh2025ladder,gannamaneni2025detecting,guimard2025classifier,yenamandra2023facts,chen2023hibug,eyuboglu2022domino}. While effective for surfacing spurious correlations in image classification (\cref{fig:introduction}(a)), these global formulations are insufficient for instance-level tasks like object detection and segmentation, where failures are often driven by contextual, relational, and spatially grounded visual patterns (\cref{fig:introduction}(c)).
Recent efforts have extended slice discovery to detection tasks by clustering object embeddings~\cite{yan2025vislix} 
or generating object tags~\cite{zhao2025hibug2}. However, representing slices as embedding clusters or tag combinations 
may not always capture complex compositional semantics (e.g., contextual, spatial, and similarity relationships). Furthermore, these methods often rely on image-level error metrics for evaluation (\cref{fig:introduction}(b)), which can restrict precise instance-level failure analysis.

\begin{figure*}[tb]
    \centering
    \includegraphics[width=0.99\textwidth]{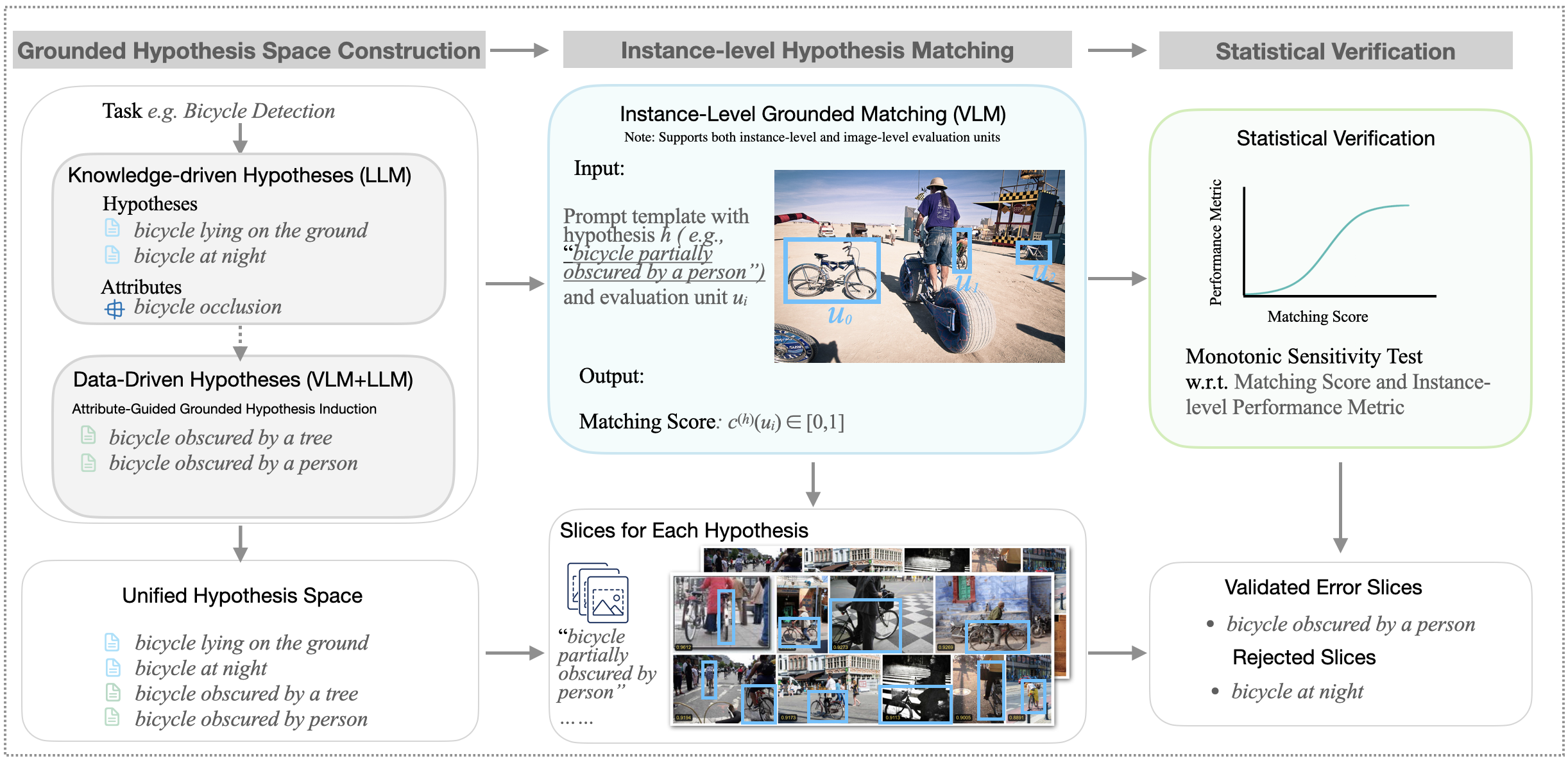}
    \caption{Overview of GH-ESD. The framework constructs grounded hypotheses and verifies them through instance-level grounding and statistical trend analysis.}
    \label{fig:slicelens}
  \end{figure*}

To address these limitations, we propose \textbf{GH-ESD} (Grounded Hypothesis-Driven Error Slice Discovery). As illustrated in \cref{fig:slicelens}, GH-ESD adopts a grounded hypothesis generate-and-verify paradigm. It integrates top-down semantic priors from a Large Language Model (LLM) with bottom-up region-level visual descriptions to construct a grounded hypothesis space of candidate failure patterns. These hypotheses are then verified using a Vision-Language Model (VLM) to ground them to specific visual regions. Crucially, to mitigate the hallucination risks of VLMs, we introduce a statistical hypothesis verification module that tests whether the model error probability increases monotonically with hypothesis intensity, thereby ensuring reliable and defensible slice identification. 

Progress in this area is further limited by the absence of principled benchmarks. 
Existing slice discovery benchmarks focus primarily on image-level classification and artificial spurious correlations (e.g., Waterbirds~\cite{sagawa2020distributionally}, NICO++~\cite{zhang2023nico++}), which do not capture the complexity of real-world instance-level failures. 

To enable standardized evaluation, we present \textbf{GESD} (Grounded Error Slice Dataset). GESD comprises expert-annotated, spatially grounded error slices derived from real-world failures of multiple object detectors and a segmentation model, establishing a ground truth for instance-level error slice discovery.

Our main contributions are:
\begin{enumerate}
    \item We propose \textbf{GH-ESD}, a grounded hypothesis generate-and-verify framework to robustly uncover systematic failure patterns. GH-ESD reframes instance-level error slice discovery as the construction of a grounded hypothesis space followed by statistical verification, enabling effective detection of spatially localized and contextual relational errors.
    \item We establish \textbf{GESD}, which is the first benchmark dedicated to instance-level error slice discovery to our knowledge. It features expert-annotated ground-truth slices, representing spatially localized and contextual relational visual difficulties.
    \item Extensive experiments demonstrate that GH-ESD consistently outperforms baselines, improving Precision@10 by 0.10 (0.73 vs. 0.63) on the GESD benchmark for detection tasks, while also supporting segmentation scenarios. We further validate the utility of discovered slices through model repair experiments, showing significant performance gains in object detection.
\end{enumerate}

\section{Related Work}
\label{sec:related_work}

\textbf{Slice Discovery as Representation or Attribute Partitioning.}
Most existing slice discovery methods implicitly treat slices as partitions in representation space or combinations of predefined attributes. 
Early approaches identify failure modes by clustering embeddings or fitting mixture models in feature space~\cite{jain2022distilling,d2022spotlight,eyuboglu2022domino,wang2023error,yenamandra2023facts}. 
For instance, Domino~\cite{eyuboglu2022domino} and FACTS~\cite{yenamandra2023facts} discover systematic errors by modeling structured subgroups in embedding space. 
While effective for image-level classification, these methods often lack interpretability, as slices correspond to regions in latent space rather than human-understandable concepts.
To improve interpretability, subsequent work adopts a ``tag-then-slice'' paradigm, first generating semantic attributes and then searching for attribute combinations associated with poor performance~\cite{chen2023hibug,luo2024llm,zhao2025hibug2,ghosh2025ladder,kim2024b2t,marani2024vigbias,guimard2025classifier}. 
Although more interpretable, this formulation assumes that failure patterns can be decomposed into independent atomic tags. 
Such attribute enumeration becomes inadequate for capturing compositional or relational visual patterns and suffers from combinatorial explosion in complex scenarios.

\textbf{Instance-Level Error Slice Discovery.}
Extending slice discovery to object detection and segmentation requires reasoning over individual instances and their spatial relationships. 
VISLIX~\cite{yan2025vislix} adopts a cluster-then-explain strategy, assisting domain experts through UI-driven ROI embedding exploration and summarizing clusters using LLMs. 
HiBug2~\cite{zhao2025hibug2} generates descriptive tags for each object instance using VLMs and searches for tag combinations correlated with errors. 
While these methods move toward instance-level clustering or bag-of-tags representations, they still rely on image-level error computation for slice evaluation, limiting their ability to perform fine-grained instance-level analysis.
In contrast, our work formulates instance-level slice discovery as grounded hypothesis space construction with statistical verification, explicitly modeling grounded patterns and probing their impact on model performance.

\textbf{Benchmarks for Slice Discovery and Robustness.}
Existing slice discovery benchmarks primarily target image-level classification and simplified spurious correlations, such as Waterbirds~\cite{sagawa2020distributionally}, CelebA~\cite{liu2015deep}, and NICO++~\cite{zhang2023nico++}. 
While useful for studying bias and subgroup robustness, these datasets do not capture the relational and spatial complexity of failures in detection and segmentation tasks. OOD benchmarks for complex tasks, including COCO-O~\cite{mao2023coco-o} and Fishyscapes~\cite{blum2021fishyscapes}, focus on distribution shift and anomaly detection rather than identifying interpretable, systematic failure slices. Consequently, they do not provide the grounded, slice-level annotations necessary for evaluating instance-level slice discovery methods. To address this gap, we introduce GESD to provide a principled evaluation testbed for instance-level slice discovery.

\section{Grounded Hypothesis-Driven Error Slice Discovery}
\label{sec:framework}
As illustrated in \cref{fig:slicelens}, we propose GH-ESD, a grounded hypothesis-driven framework that discovers instance-level error slices through hypothesis construction, matching, and statistical verification. We begin by formalizing the error slice discovery problem.
\subsection{Problem Formulation}
\label{sec:problem_formulation}

Let $\mathcal{U}=\{u_i\}_{i=1}^{N}$ denote a set of evaluation units for an instance-level vision task. 
Each evaluation unit corresponds to an object instance predicted correctly, or an error-associated region identified through prediction–ground-truth comparison, 
including (i) a ground-truth instance corresponding to a false negative (FN), 
(ii) a predicted region corresponding to a false positive (FP),
(iii) a prediction–ground-truth discrepancy region capturing localization or segmentation errors, 
or (iv) the whole image when global failure patterns are evaluated.
This unified definition allows GH-ESD to handle FNs, FPs, localization mismatches and also global errors within a single framework. Both correct and erroneous regions are evaluated to compute the performance metric for each slice.

Let $f$ be a model to be analyzed and $\ell(u_i)$ denote the performance metric (e.g., error rate, $1-\mathrm{IoU}$)
defined on evaluation unit $u_i$.

Traditional slice discovery methods define a candidate subgroup via a binary function
$g(u_i)\in\{0,1\}$, forming
$
\mathcal{S}_g=\{u_i \mid g(u_i)=1\}.
$
A subgroup is considered an error slice if
$
\mathbb{E}[\ell(u_i)\mid u_i\in\mathcal{S}_g]
-
\mathbb{E}[\ell(u_i)] > \tau,
$
for a predefined threshold $\tau$. 
Such hard partitioning relies on discrete membership decisions and is sensitive to noisy slices and fluctuations in error rates within the data.

We instead associate each hypothesis $h\in\mathcal{H}$ with a continuous matching score $c^{(h)}(u_i)\in[0,1]$,
quantifying the degree to which evaluation unit $u_i$ satisfies hypothesis $h$. 
Slice discovery is therefore reformulated as analyzing how model performance varies with respect to hypothesis intensity (quantified by the matching score):
$
\ell(u_i) \quad \text{conditioned on} \quad c^{(h)}(u_i).
$
A systematic error slice corresponds to a hypothesis for which model error increases consistently as the hypothesis is more strongly satisfied.


\subsection{Grounded Hypothesis Space Construction}
\label{sec:hypothesis_construction}

The goal of hypothesis construction is to define a hypothesis space 
$\mathcal{H}$ that systematically characterizes potential failure patterns of the model.
A hypothesis $h \in \mathcal{H}$ is defined as a semantically interpretable visual condition that can be evaluated with respect to an evaluation unit.
The hypothesis space covers both intrinsic task difficulties and dataset-specific distributional artifacts inspired by the categorization of errors in HiBug2~\cite{zhao2025hibug2}.


\textbf{Knowledge-Driven Hypotheses: Modeling Intrinsic Task Difficulty.} To capture inherent challenges of a task, we adopt a top-down hypothesis generation strategy 
guided by LLMs (prompt in supplementary Fig.~S1).
Given a concise task description, the LLM leverages world knowledge to enumerate plausible failure scenarios. 
These hypotheses typically correspond to known sources of visual task difficulty, 
such as appearance ambiguity, scale variation, object interaction, 
and challenging environmental conditions. Importantly, these hypotheses are not tags but structured semantic conditions that can later be grounded and verified.


\textbf{Data-Driven Hypotheses: Discovering Distributional Patterns.} While knowledge-driven hypotheses capture general task difficulty, 
real-world datasets often exhibit distributional biases and rare configurations 
that are not fully anticipated by prior knowledge.
To discover such patterns, we adopt a bottom-up grounded abstraction process, with prompt templates provided in supplementary Figs.~S2, S3, and~S4.
\textit{(a) Grounded Attribute-guided Caption Extraction.}
For a sampled subset of images, VLM generates 
object- or region-grounded captions based on the attributes from LLM world knowledge to avoid divergent descriptions. This attribute-guided design mitigates noise caused by unconstrained, overly diverse captions.
Unlike image-level descriptions in TCIC~\cite{kwon2024image}, these captions explicitly associate semantic attributes 
with specific visual regions, preserving spatial grounding.
\textit{(b) Attribute and Value Inference.}
An LLM analyzes the grounded captions to infer recurring semantic attributes 
and their candidate values (e.g., occlusion: by person, by car). 
This step identifies semantic dimensions present in the dataset.
\textit{(c) Hypothesis Synthesis.}
The inferred attribute–value combinations are reformulated into 
well-formed natural-language hypotheses. 
Each resulting hypothesis specifies a concrete, evaluable visual condition 
that can be directly used as input to the matching stage.

This process transforms distributional regularities into interpretable and testable hypotheses.
Knowledge-driven hypotheses provide broad coverage of intrinsic task difficulty.
In contrast, data-driven hypotheses reveal dataset-specific distributional characteristics, which may include rare or long-tail patterns that are amplified in the target data distribution due to empirical frequency and contextual biases. For the task of bicycle detection, GH-ESD may generate knowledge-driven hypotheses such as ``bicycle lying on the ground'' or ``bicycle at night'', capturing intrinsic difficulty, and data-driven hypotheses such as ``dataset-specific contextual configurations co-occurring with bicycles'', reflecting dataset-induced regularities. These hypotheses are expressed in structured natural language, 
enabling direct interaction with vision-language models for grounded verification.

\subsection{Hypothesis Matching}
\label{sec:matching}

Given a hypothesis $h\in\mathcal{H}$ and evaluation unit $u$, 
we compute a continuous matching score $c^{(h)}(u)\in[0,1]$, estimating the probability that the visual evidence associated with $u$ satisfies $h$.

A pretrained VLM capable of grounding reasoning (e.g., Qwen2.5-VL~\cite{Qwen2.5-VL}) 
is prompted with the hypothesis together with grounding signals 
(e.g., bounding boxes or point coordinates), and then produces token-level logits over its vocabulary. 
We extract logits for affirmative and negative tokens (e.g., `yes'/`no') to compute a continuous score. Prompt templates are provided in supplementary Figs.~S5 and~S6.

Given prompt $\mathcal{P}(h,u)$, let $z(t \mid \mathcal{P}(h,u))$ denote the logit corresponding to token $t$. 
We compute the matching score as

\[
c^{(h)}(u)
=
\frac{\exp\left(z(\text{yes}\mid\mathcal{P}(h,u))\right)}
{\exp\left(z(\text{yes}\mid\mathcal{P}(h,u))\right)
+
\exp\left(z(\text{no}\mid\mathcal{P}(h,u))\right)}.
\]

This probabilistic formulation converts the VLM's discriminative response into a normalized continuous score in $[0,1]$. 


\subsection{Statistical Hypothesis Verification}
\label{sec:verification}

Existing error slice discovery approaches often identify slices by thresholding empirical error rates within predefined subgroups. However, error rates can vary substantially across patterns, making fixed thresholds difficult to choose and often unreliable. Moreover, a high subgroup error rate does not necessarily indicate a systematic failure mode. Such patterns may arise from noisy grouping, weak grounding, or hallucinated hypothesis matches. Bayesian correction~\cite{yenamandra2023facts} relies on explicitly modeled uncertainty over a predefined or stable tag space, which is incompatible with our open-ended hypothesis generation framework.
We instead formulate hypothesis verification as a monotonic performance sensitivity test. 
Given a hypothesis $h$ and its matching scores $\{c^{(h)}(u_i)\}_{i=1}^{N}$ over evaluation units $\{u_i\}_{i=1}^{N}$, 
we examine whether model error probability increases monotonically with hypothesis intensity.

The matching score $c^{(h)}(u)$ reflects the degree to which evaluation unit $u$ satisfies hypothesis $h$. 
If $h$ genuinely characterizes a failure mode of model $f$, units that more strongly satisfy $h$ should exhibit higher error likelihood. Formally, we test whether the model performance metric shows a positive monotonic trend with respect to $c^{(h)}(u)$.

Low-confidence matches include irrelevant units due to matching intensity or VLM hallucination. 
Therefore, we restrict analysis to the relatively high-confidence region $\mathcal{U}_\gamma = \{u_i \mid c^{(h)}(u_i) > \gamma\}$, where $\gamma$ is an easy-to-tune empirical threshold established via confidence calibration, varying across different VLMs.

Within $\mathcal{U}_\gamma$, evaluation units are sorted in descending order of $c^{(h)}(u_i)$. 
We apply a sliding window over the ordered units. Windows containing fewer than a minimum number of units 
(e.g., $<5$) are discarded to ensure stability.

For each valid window $w$, we compute a local linear regression between matching score and performance metric:
\[
\text{slope}_w
=
\operatorname{Slope}\big(c^{(h)}(u), \ell(u)\big)_{u\in w}.
\]
The overall trend statistic for hypothesis $h$ is defined as
$
\text{Trend}(h)
=
\max_{w\in\mathcal{W}}
\text{slope}_w,
$
where $\mathcal{W}$ denotes the set of valid windows.
A hypothesis is retained as a systematic error slice if $\text{Trend}(h) > \tau_{\text{trend}}$, indicating a consistent monotonic increase in error probability with hypothesis intensity.
Unless otherwise specified, we fix $\gamma=0.5$ and $\tau_{\text{trend}}=0.2$. As shown in \cref{sec:exp_trend_vs_threshold}, the method remains stable across a broad range of parameter settings.

The max sliding-window slope is lightweight and interpretable, and it detects failure patterns that manifest only when the hypothesis is strongly satisfied and mitigates spurious matches since hallucinated or weakly grounded units do not produce a stable positive slope. The underlying monotonic sensitivity principle can also be implemented using other monotonic association measures. We additionally instantiate a multi-scale Spearman rank-correlation test, retaining a hypothesis only when the association between matching score and error is both strong and statistically significant ($\rho\!\ge\!0.5$, $p\!<\!0.05$). Both the slope- and correlation-based tests are valid realizations of the monotonic strategy; we compare them quantitatively in \cref{sec:exp_trend_vs_threshold}.
\section{{GESD} (Grounded Error Slice Dataset)}
\label{sec:dataset}

To address the limitations of existing benchmarks, we constructed GESD, a new benchmark dataset designed to capture realistic instance-level error slices.
The GESD dataset is built upon a combination of over 12K images from three widely used public validation sets: COCO (5K images)~\cite{Lin2014COCO}, KITTI (3,769 images)~\cite{Geiger2012KITTI}, and a public face detection set (3,347 images)~\cite{elmenshawii2025face}. We followed a four-step pipeline to construct error slices and ensure the quality:

\begin{enumerate}
\item \textbf{Aggregated Model Errors Collection:} We ran four representative object detectors (YOLO~\cite{wang2023yolov7}, Faster R-CNN~\cite{Ren2015FasterRCNN}, RetinaNet~\cite{Lin2017RetinaNet} and DETR~\cite{detr_2020_ECCV}) and a segmentation model (Mask R-CNN~\cite{He2017MaskRCNN}) on the combined validation images. Instead of relying on a single model, we aggregated errors from multiple models to ensure a diverse and abundant collection of failure patterns. Specifically, we collected all false positive and false negative predictions from these models, forming a pool of candidate error instances.
\item \textbf{Expert-Driven Hypotheses Definition:} Based on the aggregated error pool, three senior computer vision researchers (each with over 4 years of experience) analyzed the failure patterns. They collaboratively defined a comprehensive and representative set of error slices by formulating hypotheses. Through iterative cross-validation, they identified 21 distinct hypotheses for detection and 21 for segmentation (total 42), categorizing them into semantic confusion, contextual interference, and intrinsic visual difficulties. This expert-driven process ensures that the defined slices correspond to meaningful and systematic failure modes rather than random noise.
\item \textbf{Hypotheses Matching by Human Annotation:} This step involved rigorous human annotation to match error instances to the defined hypotheses. A team of annotators, investing a total of 1500 person-hours, reviewed every relevant instance in the dataset. For each defined hypothesis, annotators (a) verified that the instance strictly met the slice definition and exhibited the correct failure type, and (b) corrected its ground-truth bounding box and/or segmentation mask to ensure geometric accuracy.
\item \textbf{Dataset Refinement:} To create a clean ground truth for evaluation, we performed a final refinement step. For all instances not belonging to any defined error slice, model predictions were overwritten by ground truth. This creates a synthetic ``perfect'' annotation for non-error parts and ensures evaluation focuses purely on predefined slices rather than open-world unknown errors.
\end{enumerate}

The error slices in GESD contain location and category errors, including both false negatives (FN) and false positives (FP). We provide error regions with bounding boxes or segmentation masks, along with hypotheses or slice descriptions. Detailed slice statistics and error rates are provided in supplementary Sec.~S2, and the full hypothesis list is available in supplementary Tabs.~S3 and~S4.
\section{Experiments}
\label{sec:experiments}

We evaluate GH-ESD from four perspectives:
(1) effectiveness on instance-level slice discovery,
(2) ablation study of major components,
(3) generalization to classical image-level bias benchmarks, and
(4) practical utility via model repair.

We compare GH-ESD against state-of-the-art slice discovery methods, including:
(1) embedding-based methods: FD (Feature Discovery)~\cite{jain2022distilling},
Domino~\cite{eyuboglu2022domino}, and FACTS~\cite{yenamandra2023facts};
(2) interpretability-based methods: LADDER~\cite{ghosh2025ladder}, 
ViG-FACTS~\cite{marani2024vigbias}, B2T (Bias-to-Text)~\cite{kim2024b2t}, 
and ViG-B2T~\cite{marani2024vigbias}.
These approaches were originally designed for image-level bias analysis and do not explicitly or intrinsically support instance-level error slice discovery. To evaluate the proposed instance-level framework, we adapt FACTS~\cite{yenamandra2023facts} by converting bounding box predictions to image-level labels—if any bounding box in an image has an error, the entire image is marked as erroneous. HiBug~\cite{chen2023hibug} and HiBug2~\cite{zhao2025hibug2} aggregate instance errors to image-level and can be adapted to extract tags for objects. We therefore compare the adapted FACTS*, HiBug* and HiBug2 on GESD. HiBug and HiBug2 originally used GPT-3.5-Turbo and GPT-4o. We reproduced them with GPT-5.2 to avoid underestimating these baselines, while preserving their original algorithms.

Our GH-ESD implementation employs Gemini-2.5-Pro as LLM and Qwen2.5-VL-7B as VLM, selected for a trade-off between performance and efficiency (the influence of selecting different models is evaluated in supplementary Sec.~S3.1). Prompts of GH-ESD are provided in supplementary Sec.~S1.1.

Error slices can be described at different levels of granularity, so predicted and ground-truth slices may split or merge and do not correspond one-to-one. Consequently, recall becomes sensitive to the slice matching criteria, and prior work~\cite{yenamandra2023facts,eyuboglu2022domino} primarily evaluates slice discovery using Precision@k. This metric measures the precision of the top-k retrieved samples for each ground-truth error slice. Formally, for a ground truth slice $S_{gt}$ and predicted slice $S_{pred}$, Precision@k is defined as:

\begin{equation}
\text{Precision@k} = \frac{|S_{gt} \cap S_{pred}^{(k)}|}{k}
\end{equation}

where $S_{pred}^{(k)}$ represents the top-k samples in the slice ranked by hypothesis matching scores. Following prior work, we set $k=10$ by default, or the actual size if the slice has fewer than 10 samples.

The \textbf{Precision@k} evaluates accuracy of instances belonging to slices without considering hypothesis effectiveness; thus, the hypothesis precision and recall will be provided in \cref{sec:exp_relational_vs_tags}.

\subsection{Results on GESD Instance-Level Benchmark}
\label{subsec:fesd_results}

We first evaluate GH-ESD on GESD and compare against baselines: FACTS* (adapted via image-level aggregation), HiBug* (extended to instance-level tagging) and HiBug2. For the tag-based baselines, we reproduce their tag extraction procedures. HiBug* tags are interactively supplemented with instance-level attributes derived from detected objects (supplementary Sec.S3.2), while HiBug2 attributes are extracted using the released implementation. 

\begin{table}[tb]
  \centering
  \small
  \caption{\textbf{Precision@k results on GESD dataset.}  
  }
  \label{tab:fesd_results_llm}
  \setlength{\tabcolsep}{6pt}
  \begin{tabular}{lccccc}
    \toprule
    \textbf{Metric} & \textbf{FACTS*} & \textbf{HiBug*} & \textbf{HiBug2 (GPT)} & \textbf{HiBug2 (Gemini)} & \textbf{GH-ESD} \\
    \midrule
    P@5  & 0.35 & 0.34 & 0.73 & 0.77 & \textbf{0.78} \\
    P@10 & 0.28 & 0.31 & 0.63 & 0.63 & \textbf{0.73} \\
    P@20 & 0.20 & 0.26 & 0.53 & 0.52 & \textbf{0.66} \\
    \bottomrule
  \end{tabular}
\end{table}

\Cref{tab:fesd_results_llm} reports the Precision@k results on the GESD detection-based dataset. On Precision@10, the metric most commonly adopted in prior slice-discovery work~\cite{yenamandra2023facts,eyuboglu2022domino}, GH-ESD achieves 0.73, strongly outperforming HiBug2 (0.63), HiBug* (0.31) and FACTS* (0.28)—an absolute improvement of 0.10 over the strongest baseline—and it leads across all metrics, with the largest gains at P@10 and P@20. For a fair comparison, we further evaluated HiBug2 with the same Gemini-2.5-Pro backend used by GH-ESD; the results indicate that the performance gains stem from our instance-level framework rather than the choice of LLM. These improvements are attributed to the accurately described hypotheses, grounding capability, and statistical hypothesis verification of our instance-level framework.

Broken down by failure mode, GH-ESD also demonstrates superior robustness: it achieves a P@10 of 0.64 for false positives (FPs) and 0.79 for false negatives (FNs), compared to HiBug2's 0.51 and 0.72, respectively. This confirms that attribute-based tagging struggles with relational context, especially for context-heavy errors, whereas our continuous VLM reasoning over coherent natural-language hypotheses handles both FPs and FNs effectively.

\begin{figure}[tb]
  \centering
  \includegraphics[width=0.65\linewidth]{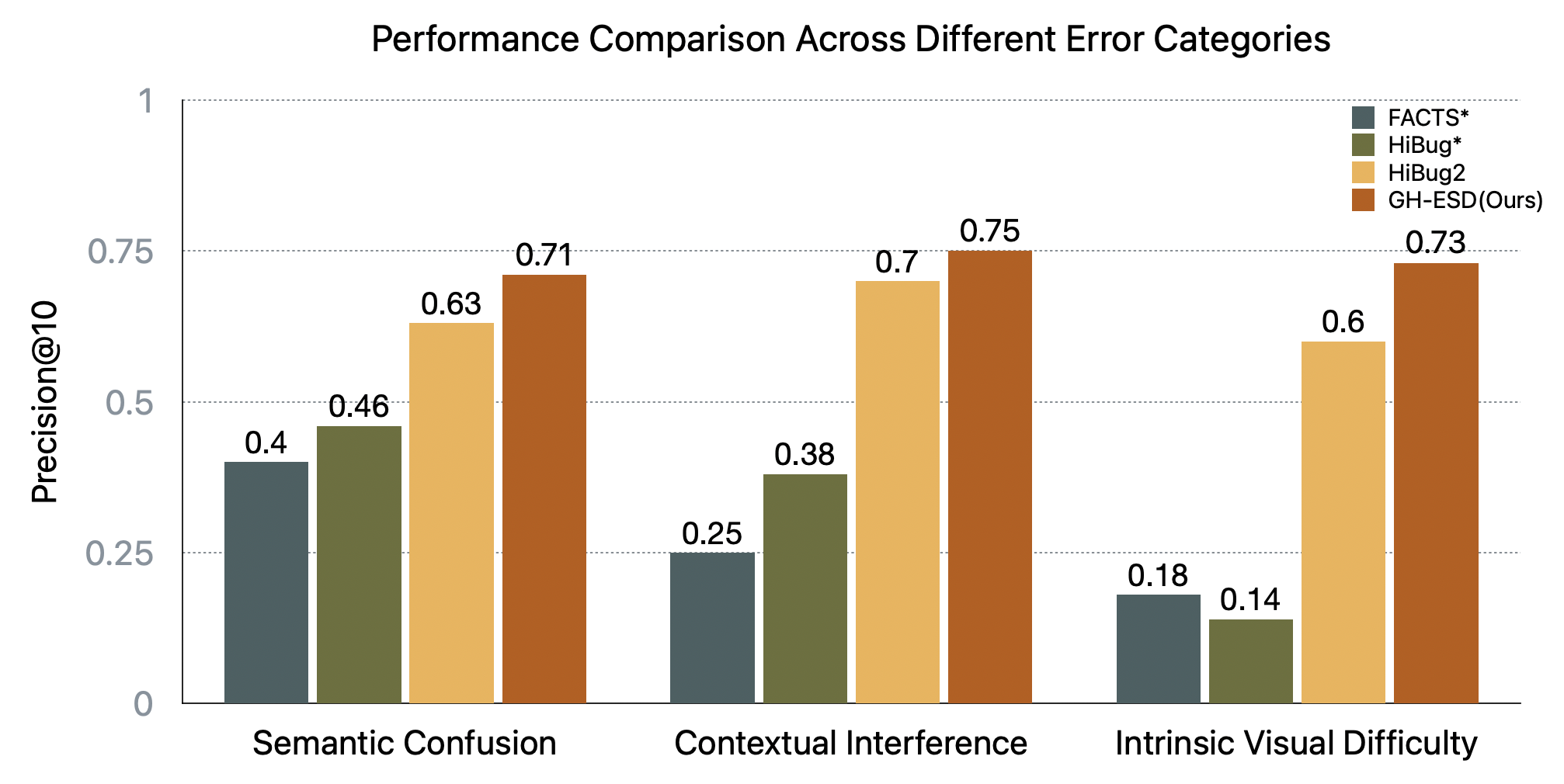}
  \caption{Performance comparison across different error categories.}
  \label{fig:category_performance}
\end{figure}

To further analyze performance across error categories, \cref{fig:category_performance} presents category-wise results under our slice taxonomy: (1) Semantic Confusion, covering errors arising from limitations in the model's conceptual understanding; (2) Contextual Interference, encompassing errors caused by an object's surroundings or relationships; and (3) Intrinsic Visual Difficulty, referring to failures caused by the inherent visual properties of the object instance itself. GH-ESD consistently outperforms baselines across all categories.

\begin{figure}[t]
  \centering
  \includegraphics[width=0.85\linewidth]{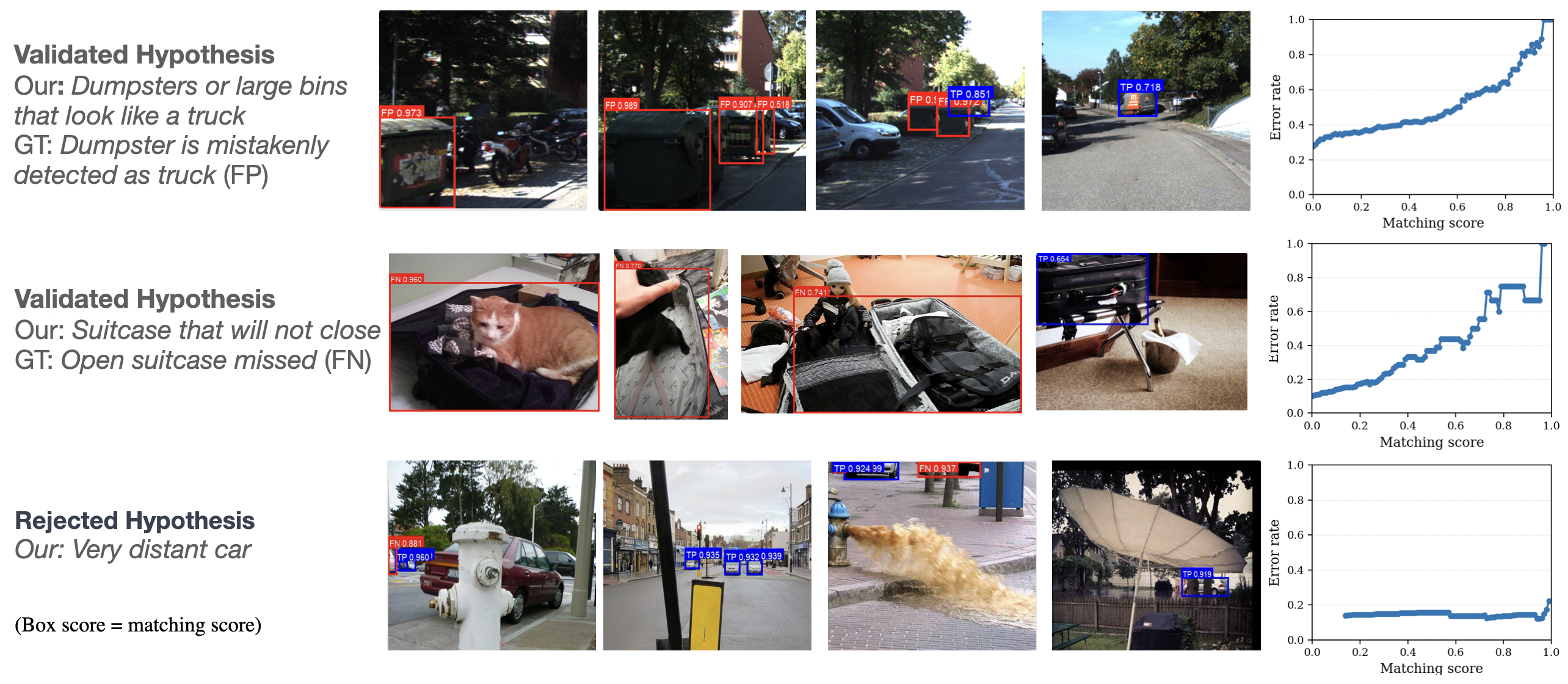}
  \caption{Examples of candidate hypotheses and their verification outcomes. The right column shows the statistical trend between matching score and error rate for each hypothesis.}
  \label{fig:qualitative_analysis}
\end{figure}

\Cref{fig:qualitative_analysis} shows several candidate hypotheses and their verification outcomes. Hypotheses such as dumpsters that visually resemble trucks and suitcases that will not close correspond to clear systematic error patterns that consistently lead to FPs or FNs. In contrast, the hypothesis of very distant cars exhibits inconsistent error trends and therefore does not form a reliable slice. Other hypotheses, such as cars occluded severely by trees, lack sufficient supporting samples, making their verification inconclusive. These examples demonstrate that our framework not only discovers meaningful error slices but also filters out spurious or statistically unsupported hypotheses.

Segmentation-based slice discovery results are reported in supplementary Tab.~S4, demonstrating that GH-ESD generalizes across both detection and segmentation tasks. The bounding box of the erroneous pixel region is used as the evaluation unit, and these results serve as baseline references for future studies.

\subsection{Ablation Study of GH-ESD}
\label{sec:exp_ablation}

We now analyze the contribution of each major component:
(1)  grounded hypotheses, (2) instance-level hypothesis matching, and
(3) statistical hypothesis verification.

\subsubsection{Grounded Hypotheses vs. Atomic Tags}
\label{sec:exp_relational_vs_tags}
We compare the proposed language-based hypotheses with attribute-based tags (HiBug2) in terms of their semantic alignment with the ground-truth failure slices, using results from \cref{subsec:fesd_results}. Specifically, an LLM-based (Gemini-2.5-Pro) evaluator determines whether a discovered slice is semantically related to any ground-truth slice (prompt in supplementary Sec.~S1.3). A slice is considered correct if the evaluator judges it as relevant. \emph{Recall} measures the proportion of ground-truth slices that are successfully retrieved, while \emph{precision} measures the proportion of discovered slices that are semantically aligned with the ground truth.

\begin{figure}[t]
  \centering
  \includegraphics[width=0.75\linewidth]{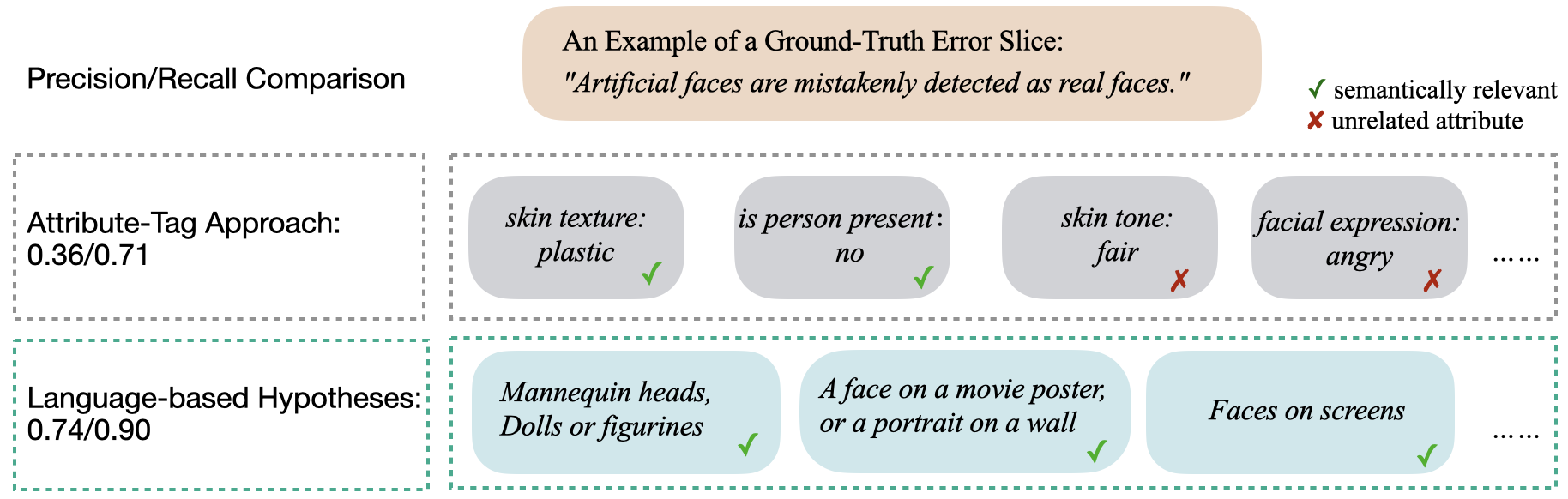}
  \caption{Comparison between tags and our hypotheses on semantic relevance with ground truth error slices.}
  \label{fig:hypothesis_comparison}
\end{figure}

The language-based hypotheses (102 slices) achieve \emph{precision/recall} of 0.74/0.90 when evaluated for semantic relevance to ground-truth error slices. In contrast, the attribute-based tagging method HiBug2 (3,666 slices) generates a large number of attribute combinations (183/713/2,770 for single/dual/triple-attribute combinations), yielding \emph{precision/recall} of 0.36/0.71 under a loose matching criterion. \Cref{fig:hypothesis_comparison} illustrates that language-based hypotheses capture the semantic structure of real-world failure patterns more coherently and compactly, while tag combinations tend to provide general semantic descriptions that lack specific failure contexts.

This performance gap reflects structural limitations of tag-based slices in the detection benchmark. Several failure modes involve interaction- or context-dependent patterns, such as ``bicycle partially occluded by a person,'' and ``a dumpster that looks like a truck.'' These conditions cannot be faithfully represented as simple conjunctions of atomic tags (e.g., object + is occluded), since atomic tags are semantically divergent and cannot readily capture relational directionality, embedded context, or visual similarity relations. Language-based hypotheses capture these structured patterns directly.

Furthermore, Knowledge-Driven (KD) and Data-Driven (DD) hypotheses exhibit strong complementarity; further analysis of the experiments in \cref{tab:fesd_results_llm} reveals that KD-only and DD-only achieve P@10 scores of 0.62 and 0.21, respectively, yet their combination reaches 0.73 with a 14\% pattern overlap.

\subsubsection{Instance-Level Grounding vs. Image-Level Aggregation}
\label{sec:exp_grounding}
We evaluate the importance of spatial grounding by comparing instance-level conditioning with image-level aggregation on the fixed hypothesis ``faces partially hidden by objects.'' Instance-level grounding achieves Precision@10 of 0.80, much higher than 0.60 under image-level analysis. Moreover, trend-based verification yields a stronger matching score-error trend statistic under instance-level conditioning than under image-level evaluation (1.32 vs.\ 0.84), confirming that instance-level conditioning mitigates dilution effects where correct predictions on unrelated instances obscure localized failures.
The LLM and VLM are kept fixed in this experiment, suggesting that the performance gains primarily stem from the grounding mechanism.

\subsubsection{Statistical Hypothesis Verification vs. Error-Rate Thresholding}
\label{sec:exp_trend_vs_threshold}
\begin{figure}[t]
  \centering
  \includegraphics[width=0.7\linewidth]{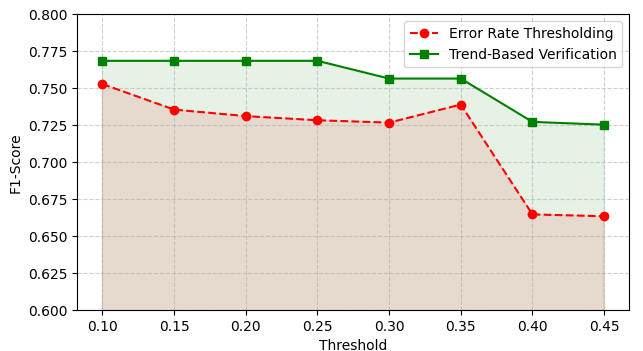}
  \caption{Comparison between error rate thresholding and statistical hypothesis verification.}
  \label{fig:correlation_comparison}
\end{figure}

We compare our monotonic trend-based hypothesis verification with traditional error-rate thresholding~\cite{ghosh2025ladder,zhao2025hibug2} using results from \cref{subsec:fesd_results}. For all discovered error slices, we compute the recall of ground-truth consistent hypotheses, the precision of validated error slices with respect to the ground truth, and report the corresponding F1 score. As shown in \cref{fig:correlation_comparison}, trend-based verification achieves an average 3.7\% improvement in F1 score and demonstrates greater robustness to threshold variation.
We further evaluate the multi-scale Spearman variant defined in \cref{sec:verification}, which achieves an F1 score of 0.78 compared to 0.77 using the original slope, and remains stable under Benjamini--Hochberg FDR correction for multiple-hypothesis control. These suggest that the effectiveness of our approach arises from the monotonic sensitivity principle itself.

\subsection{Generalization to Image-Level Bias Benchmarks}
\label{sec:exp_image_level}

Although GH-ESD is designed for instance-level reasoning, we evaluate our method on three bias discovery benchmarks: (1) Waterbirds~\cite{sagawa2020distributionally}, which contains two error-prone slices used for bias evaluation: waterbirds on land and landbirds on water; (2) CelebA~\cite{liu2015deep}, where, following~\cite{yenamandra2023facts}, we focus on the blonde hair classification task that exhibits a spurious correlation between gender and hair color; and (3) NICO++~\cite{zhang2023nico++}, where we simulate controlled spurious-correlation settings following FACTS~\cite{yenamandra2023facts} and evaluate under three correlation strengths—75\%, 90\%, and 95\%—representing increasing levels of spurious association between concepts and contexts.
Results are summarized in \cref{tab:main_results}. GH-ESD achieves competitive or state-of-the-art Precision@10 across datasets, and performs particularly well under complex spurious correlation settings on NICO++. Implementation details are provided in supplementary Sec.~S3.4.

\begin{table}[t]
  \centering
  \small
  \caption{\textbf{Precision@10 results on bias discovery datasets.} Best results are shown in \textbf{bold}, second-best are \underline{underlined}.} 
  \label{tab:main_results}
  \setlength{\tabcolsep}{6pt}
  \resizebox{\columnwidth}{!}{
  \begin{tabular}{lccc}
    \toprule
    \textbf{Method} & \textbf{Waterbirds} & \textbf{CelebA} & \textbf{NICO++(75/90/95)} \\
    \midrule
    FD & 0.90 & 0.70 & 0.19/0.19/0.19 \\
    Domino & \textbf{1.00} & \underline{0.90} & 0.24/0.25/0.27 \\
    FACTS & \textbf{1.00} & \underline{0.90} & 0.56/0.60/0.62 \\
    LADDER & 0.86 & 0.85 & 0.32/0.52/0.54 \\
    ViG-FACTS & \textbf{1.00} & \textbf{1.00} & \underline{0.60}/ \textbf{0.67}/\underline{0.65} \\
    B2T & 0.92 & 0.64 & -\\
    ViG-B2T & \underline{0.97} & 0.70 & - \\
    \midrule
    \textbf{GH-ESD (Ours)} & \textbf{1.00} & \textbf{1.00} & \textbf{0.64}/\underline{0.66}/\textbf{0.69} \\
    \bottomrule
  \end{tabular}
  }
\end{table}

\subsection{Model Repair and Practical Impact}
\label{sec:exp_repair}

\textbf{Repair on Object Detection Benchmarks.}
We evaluate whether GH-ESD can facilitate targeted model repair for object detection (e.g., bicycles). 
After verifying the error slice \emph{``bicycle partially occluded by a person''} on GESD, 
we transfer the discovered hypothesis to the COCO training set and compute instance-level slice matching scores using GH-ESD. As a baseline, we fine-tune a pre-trained Faster R-CNN on all COCO training images containing bicycle instances.
For slice-aware repair, we adapt GroupDRO~\cite{sagawa2020distributionally} to emphasize failure-prone samples identified by GH-ESD.
Specifically, images with high slice confidence are upweighted during fine-tuning, and reweighting is applied to the detection head losses. Details are in supplementary Sec.~S3.5.

As shown in \cref{tab:fg_repair}, slice-aware repair significantly improves bicycle-class mAP and mAR over the baseline. These results demonstrate that grounded, instance-level slice discovery enables more effective model adaptation.

\begin{table}[t]
\centering
\small
\caption{Model repair performance on bicycle detection. Results are averaged over 3 repetitions.}
\label{tab:fg_repair}
\setlength{\tabcolsep}{6pt}
\begin{tabular}{lcc}
\toprule
\textbf{Method} & \textbf{mAP-bicycle} & \textbf{mAR-bicycle} \\
\midrule
No model repair & 27.20   & 39.33 \\
    Baseline model repair & 28.58 $\pm$ 0.016 & 41.14 $\pm$ 0.031 \\
    \midrule
    GroupDRO + \textbf{GH-ESD} & \textbf{33.59} $\pm$ 0.028 & \textbf{47.09} $\pm$ 0.230 \\
    \bottomrule
  \end{tabular}
\end{table}

\textbf{Repair on Classification Benchmarks.}
We evaluate slice actionability on CelebA and Waterbirds following~\cite{ghosh2025ladder}. Due to fundamental differences in slice definition strategies, we employ two repair protocols. For LADDER, which produces global slice definitions, we adopt its native strategy of training a classifier per slice and ensembling their predictions. For FACTS, Domino, and GH-ESD, whose slices are defined over data subsets, we instead apply a unified DFR-based repair framework~\cite{polinadfr}. Specifically, we compute a per-sample failure score as the maximum similarity over discovered slices, and fine-tune the classifier head on high- and low-scoring subsets. Details are in supplementary Sec.~S3.5.

As shown in \cref{tab:wga_results}, GH-ESD achieves comparable or better performance than existing slice discovery baselines, indicating that the discovered slices effectively mitigate errors induced by spurious correlations in image classification.

\begin{table}[t]
  \centering
  \small
  \caption{\textbf{Worst Group Accuracy (WGA) results.} Best results are shown in \textbf{bold}, second-best are \underline{underlined}. All results are averaged over 10 independent runs to account for the stochasticity of SGD during fine-tuning.} 
  \label{tab:wga_results}
  \setlength{\tabcolsep}{6pt}
  \begin{tabular}{lcc}
    \toprule
    \textbf{Method} & \textbf{Waterbirds} & \textbf{CelebA}  \\
    \midrule
    LADDER & \underline{0.896} $\pm$ 0.014 & \underline{0.878} $\pm$ 0.015 \\
    Domino & 0.898 $\pm$ 0.022 & 0.827 $\pm$ 0.002 \\
    FACTS & 0.889 $\pm$ 0.028 & 0.600 $\pm$ 0.011\\
    \midrule
    \textbf{GH-ESD (Ours)} & \textbf{0.904} $\pm$ 0.017 & \textbf{0.882} $\pm$ 0.004 \\
    \bottomrule
  \end{tabular}
\end{table}

\section{Conclusion}
\label{sec:conclusion}

We presented \textbf{GH-ESD}, a spatially grounded, hypothesis-driven framework for instance-level error slice discovery, and introduced \textbf{GESD}, the expert-annotated benchmark enabling evaluation of instance-level slice discovery.

\textbf{Discussion and Future Work.}
Error slice discovery is inherently constrained by the coverage and bias of the evaluation dataset. When certain failure modes are underrepresented, purely statistical analysis may be insufficient to reveal them. Moreover, the hypothesis space in real-world deployments is fundamentally open-ended, and automatically generated hypotheses may not guarantee completeness.
These observations suggest that scalable error slice discovery will likely require tighter integration between automated discovery and human reasoning. Promising directions include dynamically incorporating domain-specific knowledge sources, developing adaptive hypothesis expansion mechanisms, and exploring more principled human-in-the-loop strategies for interactive debugging. Our interactive visualization tool, presented in supplementary Sec.~S4, provides an initial step toward this direction.

%
%

%
%
\clearpage
\setcounter{section}{0}
\setcounter{figure}{0}
\setcounter{table}{0}
\renewcommand{\thesection}{S\arabic{section}}
\renewcommand{\thefigure}{S\arabic{figure}}
\renewcommand{\thetable}{S\arabic{table}}
\renewcommand{\theHsection}{S\arabic{section}}
\renewcommand{\theHfigure}{S\arabic{figure}}
\renewcommand{\theHtable}{S\arabic{table}}

\title{GH-ESD: Grounded Hypothesis-Driven Error Slice Discovery for Instance-Level Vision Tasks\texorpdfstring{\\[2pt]}{ }--- Supplementary Material ---}
\titlerunning{GH-ESD --- Supplementary Material}

\author{Wei Zhang\inst{1} \and
Chaoqun Wang\inst{1} \and
Zixuan Guan\inst{1} \and
PingSheng Kao\inst{2} \and
Pengfei Zhao\inst{1} \and
Peng Wu\inst{1} \and
Sifeng He\inst{1}}

\authorrunning{W. Zhang et al.}

\institute{Apple, Beijing, China \and Apple, USA}

\maketitle

\section{Prompts}
\subsection{System Prompt Templates Used in GH-ESD}
\label{subsection:prompts_slicelens}
A complete prompt consists of a system prompt template and a task-specific user prompt. For specific tasks, task descriptions are provided in user prompts, as detailed in \cref{subsec:dataset_prompts}. This section presents detailed system prompt templates used in GH-ESD hypothesis generation module.
\subsubsection{Knowledge-driven Hypothesis Generation Prompt}
The system prompt guides the model to consider various factors, such as object attributes and image quality, and to structure its output in a specific JSON format. This structured output contains task-specific difficulties and attributes that affect model performance. The task-specific difficulties serve as language-based hypotheses for slice discovery (the `search' type in the prompt), while the attributes are used for data-driven hypothesis generation (the `cluster' type in the prompt). Details are shown in \cref{fig:appendix_hypothesis_prompt}.

\begin{figure*}[t]
\centering
\begin{tcolorbox}[
    colback=orange!5!white,
    colframe=orange!60!black,
    title=\textbf{System Prompt for Knowledge-driven Hypothesis Generation},
    fonttitle=\bfseries\small,
    arc=2mm,
    boxrule=0.5pt,
    width=\textwidth
]
\scriptsize
You are an expert in computer vision failure analysis. A user will provide the \texttt{\{Task Description\}} (e.g., "a cat body detection"). Your goal is to identify and list all possible reasons why a computer vision model might fail on this task. In your analysis, consider a comprehensive set of factors, including but NOT LIMITED TO:

\begin{itemize}[leftmargin=*, topsep=0pt, itemsep=0pt, parsep=0pt]
    \item Object Attributes and interactions
    \item Background Descriptions
    \item Image Qualities
    \item Environmental Conditions
    \item Limitations in handling edge cases or extreme conditions
    \item Camera capturing conditions
    \item Any additional domain-specific challenges
\end{itemize}

For the given task description, generate a structured and detailed list of failure hypotheses. Your output must include:
\begin{itemize}[leftmargin=*, topsep=0pt, itemsep=0pt, parsep=0pt]
    \item Comprehensive different factors (e.g., Object Attributes, Image Qualities, etc.).
    \item Under each factor, list comprehensive failure patterns.
    \item For each failure pattern, include:
    \begin{itemize}[topsep=0pt, itemsep=0pt, parsep=0pt]
        \item A title that summarizes the potential failure reason.
        \item A description that explains in detail why this might be an issue.
        \item A list of prompts to further explore the hypothesis. Each prompt must include:
        \begin{itemize}[topsep=0pt, itemsep=0pt, parsep=0pt]
            \item A prompt field with a specific query or cluster description.
            \item A type field, where the value is either ``cluster'' or ``search''. Use ``cluster'' if the prompt provides a broad description with many possible attributes for further investigation (e.g., ``cat colors'' or ``cat types''). Use ``search'' if the prompt can be directly used to retrieve relevant images (e.g., ``a white cat'').
        \end{itemize}
    \end{itemize}
    \item \textbf{IMPORTANT:} Ensure that your final output does not contain any duplicate failure patterns or prompts. Every entry must be unique and directly relevant to image retrieval or image clustering.
\end{itemize}

Please think through at least three iterations internally to ensure completeness, accuracy, and uniqueness. Then, output your final results in a valid JSON object formatted exactly as follows:

\begin{lstlisting}[language=json, basicstyle=\ttfamily\tiny, breaklines=true]
{
  "title": "Possible failure reasons for [task description] model",
  "hypothesis": {
    "Object Attributes": [
      {
        "title": "[Short title for failure pattern 1]",
        "description": "[Detailed explanation]",
        "prompts": [
          {
            "prompt": "[related query or cluster description term]",
            "type": "[cluster or search]"
          },
          ...
        ]
      },
      ...
    ],
    "Background Descriptions": [ ... ],
    ... (other categories as comprehensively as possible)
  }
}
\end{lstlisting}
\end{tcolorbox}
\caption{The detailed system prompt for generating a comprehensive set of knowledge-driven hypotheses. The \texttt{\{Task Description\}} placeholder is replaced with the
specific task being analyzed.
}
\label{fig:appendix_hypothesis_prompt}
\end{figure*}

\subsubsection{Data-driven Hypothesis Generation Prompt}
\begin{itemize}
\item{\textbf{Grounded Attribute-aware Caption Extraction Prompt}}. This prompt directs the model to focus on a specific image region defined by bounding box coordinates and to generate a descriptive caption that highlights potential reasons for failure related to a given attribute. The attribute originates from the ``cluster'' type prompts discussed in the preceding section. This process is crucial for generating fine-grained, instance-level descriptions from datasets. (Details are shown in \cref{fig:appendix_caption_prompt}).

\begin{figure*}[t]
\centering
\begin{tcolorbox}[
    colback=orange!5!white,
    colframe=orange!60!black,
    title=\textbf{Grounded Attribute-aware Caption Extraction},
    fonttitle=\bfseries\small,
    arc=2mm,
    boxrule=0.5pt,
    width=\textwidth
]
\scriptsize
You are an AI model designed to analyze images and provide detailed captions for a specific region relevant to the \texttt{\{Attribute\}}. The specified region is defined by bounding box coordinates in the format \texttt{[xmin, ymin, xmax, ymax]}.

\end{tcolorbox}
\caption{The prompt used to instruct the Vision-Language Model (VLM) for grounded attribute-aware caption extraction. The \texttt{\{Attribute\}} placeholder is replaced with the specific attribute being analyzed.}
\label{fig:appendix_caption_prompt}
\end{figure*}

\item{\textbf{Attribute Inference Prompt}}. This section outlines a two-step process for inferring attribute values from captions. In the first step, a prompt instructs a Large Language Model (LLM) to extract values or keywords from captions based on given attributes. The second step involves processing the raw list of extracted values or keywords by removing duplicates and irrelevant terms and organizing the remainder into semantically coherent categories. This structured process transforms unstructured outputs from the Vision-Language Model (VLM) into a clean, hierarchical set of attributes and values for subsequent data-driven hypothesis generation. (Details are shown in \cref{fig:appendix_clustering_prompt}).
\begin{figure*}[t]
\centering
\begin{tcolorbox}[
    colback=orange!5!white,
    colframe=orange!60!black,
    title=\textbf{Two-Step Prompt for Attribute Inference},
    fonttitle=\bfseries\small,
    arc=2mm,
    boxrule=0.5pt,
    width=\textwidth
]
\footnotesize

\textbf{Step 1: Attribute Value or Keyword Extraction}
\vspace{1mm}

You will be given a description of the \texttt{\{Attribute\}}. 
Your job is to determine the most relevant \texttt{\{Attribute\}} based on the provided description.
Don't return anything except the result.

\textbf{Example Input:}
\begin{verbatim}
Attribute: activity
Caption: The image depicts a person performing a handstand on a purple 
mat in a spacious, well-lit room with white walls and a high ceiling.
\end{verbatim}
\textbf{Expected Output:}
\begin{verbatim}
handstand
\end{verbatim}

If the description is irrelevant to \texttt{\{Attribute\}}, please output 'none'.

Now, process the provided list: 

\hrulefill

\vspace{2mm}

\textbf{Step 2: Keyword Refinement and Clustering}
\vspace{1mm}

You are a helpful text clustering assistant. You will be given a list of keyword text. Your job is to refine and categorize the list to clusters. Your goal is to process the provided keyword list to:
\begin{itemize}
    \item \textbf{Eliminate Duplicate Semantics:} Identify and remove keywords that represent similar meaning, even if expressed differently (e.g., 'car' and 'automobile').
    \item \textbf{Eliminate irrelevant keywords:} Identify and remove keywords that are not related to \texttt{\{Attribute\}}.
    \item \textbf{Group into Categories:} Organize keywords into distinct, non-overlapping categories based on their meaning and context.
    \item \textbf{Ensure Clarity:} Avoid overly broad or overly narrow categories. Each keyword should fit logically within its assigned category.
\end{itemize}

\textbf{Example Input:}
\begin{verbatim}
['car', 'automobile', 'vehicle', 'bike', 'bicycle', 'motorcycle', 
'transport']
\end{verbatim}
\textbf{Expected Output (JSON format):}
\begin{verbatim}
{"transport":["car", "bike", "motorcycle"]}
\end{verbatim}

Please think and improve your results at least three turns. And then eliminate duplicate semantics in one more turn.
Now, process the provided keyword list and output the final results in JSON format at the end with unique, non-overlapping semantics.

\end{tcolorbox}
\caption{The prompts used for attribute inference via text clustering. Step 1 extracts relevant values from attributes and captions, and Step 2 clusters them into semantic categories. The \texttt{\{Attribute\}} placeholders are replaced with the specific attribute being analyzed.}
\label{fig:appendix_clustering_prompt}
\end{figure*}

\item{\textbf{Refinement Prompt for Grounded Query Phrases} }. This prompt instructs the model to convert a structured list of semantic attributes and values into natural language hypotheses suitable for text-to-image retrieval. The strict constraint to preserve semantics ensures that the resulting queries accurately reflect the discovered error attributes and values, forming our data-driven hypothesis. (Details are shown in \cref{fig:appendix_refinement_prompt}).

\begin{figure*}[t]
\centering
\begin{tcolorbox}[
    colback=orange!5!white,
    colframe=orange!60!black,
    title=\textbf{Refinement Prompt for Grounded Query Phrases},
    fonttitle=\bfseries\small,
    arc=2mm,
    boxrule=0.5pt,
    width=\textwidth
]
\small
You will be given a JSON list of attributes with values or cluster names with categories. These cluster names are summarized from descriptions related to the \texttt{\{Attribute\}}.
Your job is to compose a text to image retrieval prompt from each category and item in the list. 
The prompt should be phrases or sentences not keywords and only contain the given information. 

You can start or end with general descriptions, for example ``a photo of''  etc. Please DO NOT add or modify the context and semantics. 
The output JSON format is:
\begin{lstlisting}[language=json, basicstyle=\ttfamily\scriptsize]
{"results": [prompt1, prompt2, ...]}
\end{lstlisting}

Now process the list and generate prompts one by one for each item.

\end{tcolorbox}
\caption{The prompt used for refining attributes and values into grounded query phrases. The \texttt{\{Attribute\}} placeholder is replaced with the specific task context and attribute being analyzed.}
\label{fig:appendix_refinement_prompt}
\end{figure*}

\end{itemize}

\subsubsection{Hypothesis Matching Prompt for Point-based Grounding Signal}
\begin{itemize}
\item{\textbf{Point-based prompt template}}. This prompt shown in \cref{fig:appendix_point_prompt} grounds hypothesis together with point signals based on VLM inference.
\begin{figure*}[t]
\centering
\begin{tcolorbox}[
    colback=orange!5!white,
    colframe=orange!60!black,
    title=\textbf{Point-based Prompt Template},
    fonttitle=\bfseries\small,
    arc=2mm,
    boxrule=0.5pt,
    width=\textwidth
]
\small
\texttt{"If the region pointed to by {point\_2d: $(x, y)$} matches the description \texttt{\{Hypothesis\}}, please answer yes else no."}
where $(x, y)$ denotes the 2D coordinate within the error region.
\end{tcolorbox}
\caption{The prompt for grounded hypothesis matching with point signals. The \texttt{\{Hypothesis\}} and \texttt{\{(x,y)\}} placeholders are replaced with the specific hypothesis and the point coordinates being analyzed.}
\label{fig:appendix_point_prompt}
\end{figure*}
\end{itemize}

\subsubsection{Hypothesis Matching Prompt for Bounding box-based Grounding Signal}
\begin{itemize}
\item{\textbf{Bounding box-based prompt template}}. This prompt shown in \cref{fig:appendix_bbox_prompt} grounds hypothesis together with bbox signals based on VLM inference.
\begin{figure*}[t]
\centering
\begin{tcolorbox}[
    colback=orange!5!white,
    colframe=orange!60!black,
    title=\textbf{Bounding Box-based Prompt Template},
    fonttitle=\bfseries\small,
    arc=2mm,
    boxrule=0.5pt,
    width=\textwidth
]
\small
\texttt{"If the region $[x_{\min}, y_{\min}, x_{\max}, y_{\max}]$ matches the description \texttt{\{Hypothesis\}}, please answer yes else no."} 
\end{tcolorbox}
\caption{The prompt for grounded hypothesis matching with bounding box signals. The \texttt{\{Hypothesis\}} placeholder and coordinates are replaced with the specific hypothesis and region being analyzed.}
\label{fig:appendix_bbox_prompt}
\end{figure*}
\end{itemize}

\subsection{User Prompts for Hypothesis Generation on Specific Datasets}
\label{subsec:dataset_prompts}

The following user prompts are used as task context to generate knowledge-driven hypotheses for the specific tasks or datasets in GH-ESD. The corresponding system prompts are shown in \cref{fig:appendix_hypothesis_prompt}. Each prompt is tailored to the specific characteristics and challenges of its respective dataset.

\subsubsection{User Prompts for GESD Dataset}

For the GESD (Grounded Error Slice Dataset) experiments, we use two separate prompts: one for detection-related error slices and another for segmentation-related error slices. The user prompt used for object detection hypothesis generation is shown in \cref{fig:fesd_detection_prompt}, which focuses on three main error types: false positives, false negatives, and misclassification.
\Cref{fig:fesd_segmentation_prompt} shows the comprehensive prompt for instance segmentation tasks, addressing failure modes such as boundary errors, incomplete masks, over-segmentation, and under-segmentation.

\begin{figure*}[t]
\centering
\begin{tcolorbox}[
    colback=blue!4!white,
    colframe=blue!35!black,
    title=\textbf{User Prompt for GESD Object Detection Hypothesis Generation},
    fonttitle=\bfseries\small,
    arc=2mm,
    boxrule=0.5pt,
    width=\textwidth
]
\small
An object detection model designed to detect \texttt{\{label\}}.
Specifically, consider and describe possible scenarios for the following types of errors:

\vspace{2mm}
\textbf{False Positive:} In what situations might other objects be mistakenly detected as \texttt{\{label\}}?

\vspace{2mm}
\textbf{False Negative:} In what situations might \texttt{\{label\}} fail to be detected by the model?

\vspace{2mm}
\textbf{Misclassification:} In what situations might \texttt{\{label\}} be incorrectly detected as another object class?

\end{tcolorbox}
\caption{The specialized user prompt used for object detection knowledge-driven hypothesis generation on the GESD dataset. The prompt focuses on the three main types of detection errors: false positives, false negatives, and misclassification. The \texttt{\{label\}} placeholder is replaced with the specific object class being analyzed.}
\label{fig:fesd_detection_prompt}
\end{figure*}

\begin{figure*}[t]
\centering
\begin{tcolorbox}[
    colback=blue!4!white,
    colframe=blue!35!black,
    title=\textbf{User Prompt for GESD Instance Segmentation Hypothesis Generation},
    fonttitle=\bfseries\small,
    arc=2mm,
    boxrule=0.5pt,
    width=\textwidth
]
\small
An Instance Segmentation model designed to detect and segment \texttt{\{label\}}.
Think systematically from the following perspectives, and describe possible scenarios for each type of error below:

\textbf{False Positive:} In what situations might other objects be mistakenly segmented as \texttt{\{label\}}?

\textbf{False Negative:} In what situations might \texttt{\{label\}} fail to be segmented?

\textbf{Misclassification:} In what situations might \texttt{\{label\}} be incorrectly segmented as another object class?

\textbf{Boundary Error:} When might the segmentation mask boundary deviate from the true object contour (e.g., occlusion, uneven lighting, blurry edges)?

\textbf{Incomplete Mask:} When might the model segment only part of the \texttt{\{label\}} instead of the entire object?

\textbf{Over-segmentation:} When might the model mistakenly split one \texttt{\{label\}} instance into multiple segments?

\textbf{Under-segmentation:} When might the model merge multiple nearby \texttt{\{label\}} instances into a single mask?

\end{tcolorbox}
\caption{The specialized user prompt used for instance segmentation knowledge-driven hypothesis generation on the GESD dataset. This prompt addresses comprehensive errors (false positives, false negatives, misclassification, boundary errors, incomplete masks, over-segmentation, under-segmentation) to cover potential failure patterns thoroughly in instance segmentation tasks. The \texttt{\{label\}} placeholder is replaced with the specific object class being analyzed.}
\label{fig:fesd_segmentation_prompt}
\end{figure*}

These task-specific user prompts enable GH-ESD to generate comprehensive failure hypotheses that cover all relevant aspects of detection and instance segmentation models, ensuring thorough coverage of potential error patterns in instance-level computer vision tasks.

\subsubsection{User Prompt for Waterbirds Dataset}
The user prompt for the Waterbirds Dataset is as follows:
\begin{tcolorbox}[colback=gray!10, colframe=gray!50, boxrule=0.3pt]
A model that classifies whether a bird is a water bird or a land bird. The dataset is constructed by cropping out birds from photos and transferring them onto different backgrounds.
\end{tcolorbox}
\subsubsection{User Prompt for CelebA Dataset}
The user prompt for the CelebA dataset is as follows:
\begin{tcolorbox}[colback=gray!10, colframe=gray!50, boxrule=0.3pt]
A model that detects whether a person in an image has blonde hair, considering various attributes of the person.
\end{tcolorbox}

\subsubsection{User Prompt for NICO++ Dataset}
The user prompt for the NICO++ Dataset is as follows:
\begin{tcolorbox}[colback=gray!10, colframe=gray!50, boxrule=0.3pt]
A model is used for multi-label classification, classifying images into one of six categories: mammals, birds, plants, airways, waterways, and landways. The dataset consists of real-world images of these categories appearing in six different contexts (dim lighting, outdoor, grass, rock, autumn, and water). 
\end{tcolorbox}

\subsection{Semantic Relevance Evaluator Prompt}
Each algorithm output is processed individually using the following prompt for semantic relevance evaluation:
\label{sec:evaluator_prompt}
\begin{tcolorbox}[
    colback=gray!3,
    colframe=gray!30,
    arc=1mm,
    boxrule=0.3pt,
    top=2pt,
    bottom=2pt,
    left=2pt,
    right=2pt
]
\footnotesize
I have a description of an error pattern from a computer vision model, and a list of ground truth error patterns.

Please identify which of the following ground truth patterns is semantically the same as the \{algorithm output\}.
For example, Non-upright face (nonvertical pose) is similar to faces with extreme head tilt, but not similar to bicycle lying on ground or bicycle upside down.
The ground truth patterns are provided below as a numbered list.

Please respond with ONLY the number of the matching ground truth pattern. If none of the ground truth patterns are a good semantic match, respond with ``-1''.
NOTE: The ground truth patterns and algorithm patterns should indicate the same objects, context, conditions and actions.
\end{tcolorbox}

\section{GESD Dataset Details}
\label{section:gesd_detail}

\begin{figure*}[t]
    \centering
    \includegraphics[width=\linewidth]{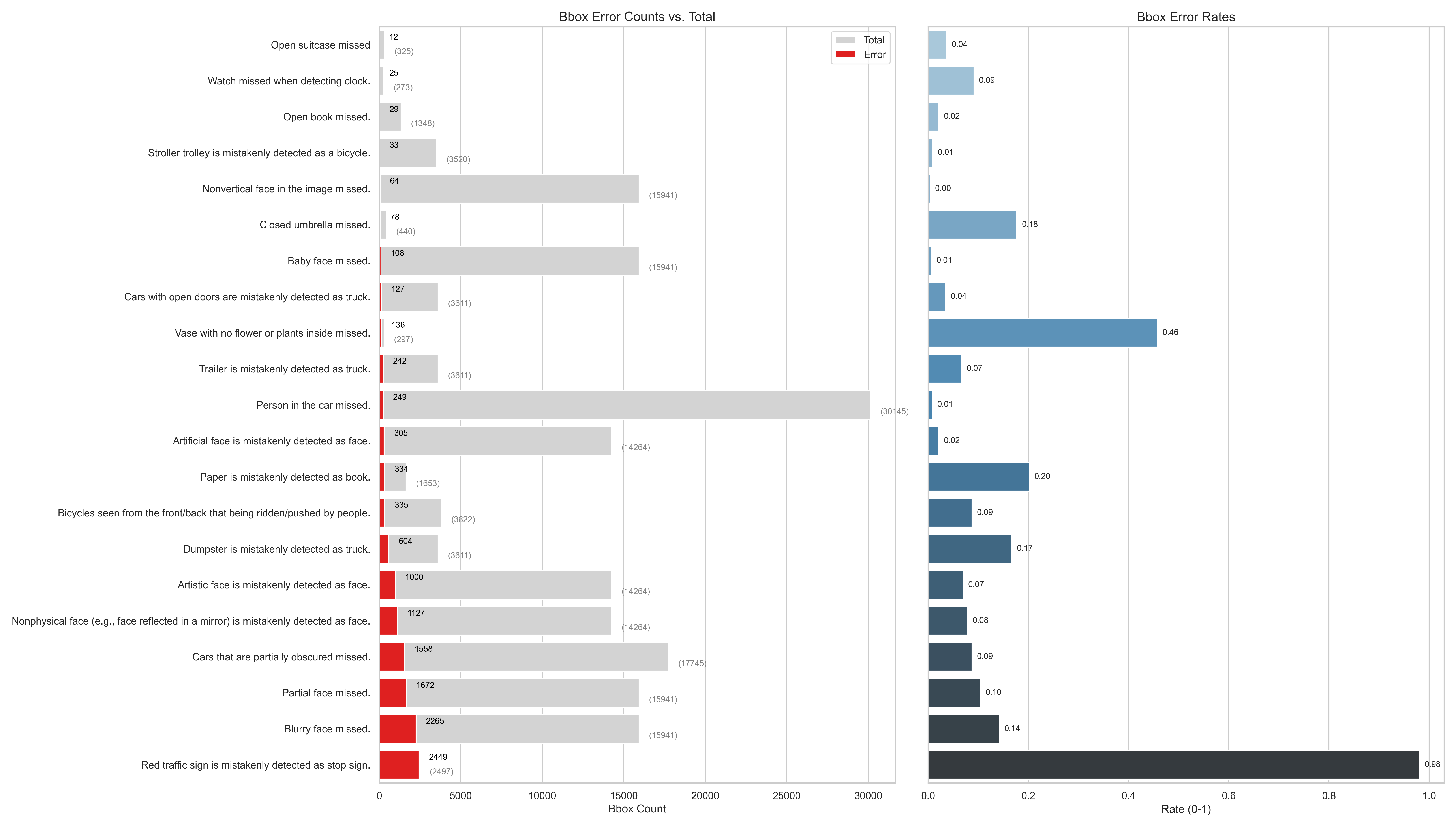}
    \caption{\textbf{Distribution of instance-level detection error statistics on the GESD dataset.}}
    \label{fig:dec_bbox_error_bar_charts}
\end{figure*}

\begin{figure*}[t]
    \centering
    \includegraphics[width=\linewidth]{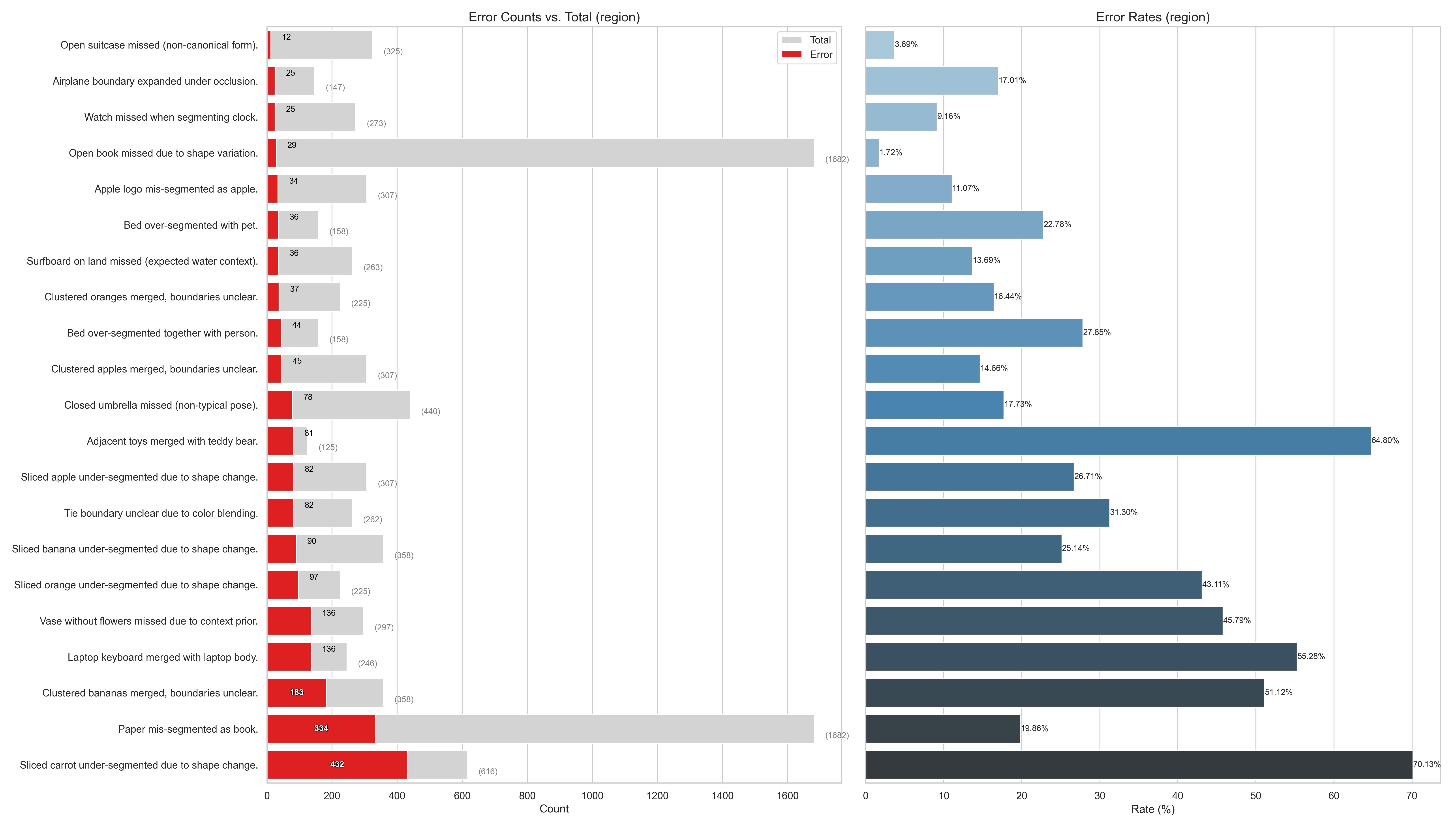}
    \caption{\textbf{Distribution of instance-level segmentation error statistics on the GESD dataset.}}
    \label{fig:seg_region_error_bar_charts}
\end{figure*}

The distribution of detection and segmentation error slices is summarized in this section. 
\Cref{fig:dec_bbox_error_bar_charts} shows detection-related slices, while \cref{fig:seg_region_error_bar_charts} shows segmentation-related slices.

We report error rates separately for False Negative (FN) and False Positive (FP) errors that are related to different slices.
FN-dominant errors (missed detections or segmentations) are computed as $\text{FN}/(\text{TP}+\text{FN})$ over ground-truth instances, while FP-dominant errors (over-segmentation, mis-segmentation, or mis-detection) are computed as $\text{FP}/(\text{TP}+\text{FP})$ over predicted instances.

The left panels show the absolute number of errors (red bars) together with the total number of relevant instances (grey bars), while the right panels show the corresponding error rates. 
These statistics reveal that some error types are frequent in absolute terms, whereas others exhibit higher conditional error rates despite smaller sample sizes.

\section{More Detailed Experimental Results}
\subsection{LLMs and VLMs Selection in GH-ESD}
\label{subsection:llm_selection}
We compare different LLMs for constructing slice hypotheses: Gemini-2.5-Pro (our choice), GPT-4o, and Claude-3.5-Sonnet. Performance variations are minimal (0.69 vs. 0.68 vs. 0.67 on NICO++ (95\%)), suggesting that the constructing method is robust to LLM choice. This robustness indicates that the world knowledge of powerful LLM in vision tasks is sufficient for analyzing task-intrinsic difficulties and data distribution attributes. The effectiveness of GH-ESD primarily depends on the vision-language grounding component rather than the specific language model used for hypothesis generation.

\begin{table}[t]
  \centering
  \small
  \caption{\textbf{Comparison of Different VLMs for Hypothesis Matching on NICO++ (95\%).} GPT-4o tested with 10 concurrent requests. SigLIP times exclude embedding extraction overhead.}
  \label{tab:vlm_comparison}
  \setlength{\tabcolsep}{6pt}
  \begin{tabular}{lcc}
    \toprule
    \textbf{VLM} & \textbf{Precision@10} & \textbf{Speed} \\
    \midrule
    SigLIP & 0.51 & 0.1 s / 1000 images \\
    GPT-4o & 0.71 & 5 min / 1000 images \\
    Qwen2.5-VL-7B & 0.69 & 10 s / 1000 images \\
    \bottomrule
  \end{tabular}
\end{table}

We evaluated three different Vision-Language Models (VLMs) for hypothesis matching to analyze the trade-off between effectiveness and computational efficiency. As shown in \cref{tab:vlm_comparison}, our evaluation on the NICO++ (95\%) dataset reveals a clear distinction between model families.

Embedding-based models like SigLIP are exceptionally fast due to their use of pre-computed embeddings, but they achieve lower precision as their global representations struggle with instance discrimination and grounded reasoning. In contrast, reasoning-based models demonstrate superior performance. While GPT-4o achieves the highest precision@10 (0.71), its API latency is prohibitive at 5 minutes per 1,000 images. 

Qwen2.5-VL-7B, deployed locally with the vLLM~\cite{S-kwon2023efficient} framework, offers the best practical balance. To optimize its performance, we structure the hypotheses matching prompts to share the same system prompt and image prefix. Leveraging prefilling technology, only the text in the user prompts requires processing after the initial hypothesis matching score computation. Furthermore, the model's output is restricted to ``yes'' or ``no'', which makes the decoding process highly efficient. As a result, it delivers competitive precision (0.69) with a significantly lower latency of 10 seconds per 1,000 images on one A100 GPU. This makes it our choice for GH-ESD.

We conduct an ablation study on logit selection of VLM to evaluate statistical robustness. We evaluate three different matching score computation strategies on a subset of the GH-ESD benchmark (mAP across 5 runs): (1) using affirmative and negative tokens (yes/no logits), which achieves 0.47; (2) using direct numerical outputs, achieving 0.23; and (3) a categorical ``yes/no/not-sure'' strategy, achieving 0.18. Direct numerical outputs suffer from VLM regression instability, while the ``yes/no/not-sure'' strategy reintroduces quantization errors analogous to binary tags. By contrast, the proposed yes/no probabilistic logit conversion provides the smoothest signal, suppressing spurious matches and facilitating more robust statistical verification.

\subsection{HiBug Baseline Implementation Details}
\label{subsec:hibug_details}

For comparison with HiBug on our GESD dataset, we generated a comprehensive attribute corpus following the HiBug methodology interactively. We generated visual attributes relevant to object detection failures by chatting with ChatGPT. \Cref{tab:hibug_attributes} presents the complete set of attributes used for HiBug's instance-level analysis on detection tasks.

\begin{table*}[t]
\centering
\scriptsize
\caption{\textbf{HiBug attribute corpus generated for instance-level error slice discovery on GESD dataset.} Attributes are generated interactively using ChatGPT following HiBug methodology. Each attribute includes categorical values, prompt templates, and evaluation questions for systematic analysis.}
\label{tab:hibug_attributes}
\setlength{\tabcolsep}{4pt}
\resizebox{\linewidth}{!}{
\begin{tabular}{llp{3cm}p{4cm}p{5cm}}
\toprule
\textbf{Attribute} & \textbf{Type} & \textbf{Values} & \textbf{Prompt Template} & \textbf{Evaluation Question} \\
\midrule
Object Reality & Description & real\_object, image\_in\_image, artistic\_rendering, reflection\_mirror, screen\_display & A detected object in object reality & What is the object in the photo, and is it a true representation of reality? \\
\midrule
Image Sharpness & Description & very\_sharp, sharp, slightly\_blurry, blurry, very\_blurry & A detected object with sharp image\_sharpness & How would you rate the sharpness of the detected object in the photo? \\
\midrule
Object Visibility & Description & 100\_percent, 80\_percent, 50\_percent, 30\_percent, less\_than\_20 & A detected object with 90\% object visibility & What is the object\_visibility\_percentage of the detected object in the photo? \\
\midrule
Vehicle Subtype & Description & car, truck, bus, trailer, van, motorcycle, bicycle, stroller, cart, dumpster & A detected car with a SUV subtype & What is the vehicle\_subtype of the detected object in the photo? \\
\midrule
Person Activity & Description & standing, walking, sitting, riding\_bicycle, in\_vehicle, running, lying & A \#DETECTED\_OBJECT doing \#1 & What is the activity of the person in the photo? \\
\midrule
Face Age Group & Description & infant, child, teenager, young\_adult, adult, elderly & A \#LABEL in the youth \#1 & What age group does the detected object's face belong to in the photo? \\
\midrule
Face Orientation & Description & upright\_frontal, upright\_side, tilted, horizontal, upside\_down & A \#LABEL with \#1 face\_orientation & What is the orientation of the face in the detected object in the photo? \\
\midrule
Object State & Description & closed, partially\_open, fully\_open, in\_use, idle & A \#LABEL in motion & What is the object\_state of the detected object in the photo? \\
\midrule
Object Material & Description & solid, paper, metal, glass, plastic, fabric & A \#1 ball with metal material & What material is the detected object in the photo made of? \\
\midrule
Object Scale & Description & fills\_frame, dominates, medium, small, tiny & A detected object with a small object scale in the frame & What is the object\_scale\_in\_frame of the detected object in this photo? \\
\midrule
Object Size & Description & small, medium, large & A \#LABEL with small object\_size & What is the size of the detected object in the photo? \\
\midrule
Occlusion Level & Description & not\_occluded, partially\_occluded, heavily\_occluded & A detected object with low occlusion level & What is the occlusion level of the detected object in the photo? \\
\midrule
Motion Blur & Binary & yes, no & - & Is the detected object in the photo showing signs of motion blur? \\
\midrule
Lighting Condition & Description & bright, normal, dark & An \#LABEL in \#1 lighting condition & What is the lighting condition of the detected object in the photo? \\
\midrule
View Angle & Description & frontal, side, back, top & A \#LABEL from a \#1 view angle & What is the view angle of the detected object in the photo? \\
\midrule
Object Truncation & Description & complete, partially\_truncated, heavily\_truncated & A truncated detected object & What is the object\_truncation value of the detected object in the photo? \\
\midrule
Color Contrast & Description & low\_contrast, medium\_contrast, high\_contrast & A \#LABEL with high color contrast & What is the color contrast of the detected object in the photo? \\
\midrule
Background Clutter & Description & simple\_background, moderate\_clutter, heavy\_clutter & A \#LABEL with busy background\_clutter & What is the background clutter of the detected object in the photo? \\
\midrule
Object Distance & Description & close, medium, far & A detected object at close distance & What is the distance of the detected object in the photo? \\
\midrule
Object Pose & Description & upright, tilted, horizontal, inverted & An \#LABEL in a \#1 pose & What is the pose of the detected object in the photo? \\
\midrule
Mirror/Glass Presence & Binary & yes, no & - & Is there a mirror or glass present in the photo? \\
\midrule
Color Similarity & Description & distinct, similar\_to\_background, blends\_with\_background & A \#LABEL with \#1 object\_color\_similarity & What is the similarity between the detected object and the color of the photo? \\
\midrule
Object Reflection & Description & no\_reflection, slight\_reflection, strong\_reflection & A detected object with a shiny reflection & What does the detected object in the photo reflect? \\
\midrule
Time of Day & Description & morning, noon, afternoon, evening, night & A sky with morning light & What is the time of day of the detected object in the photo? \\
\midrule
Texture Pattern & Description & smooth, textured, patterned & A \#LABEL with \#1 texture pattern & What is the texture pattern of the detected object in the photo? \\
\midrule
Partial Visibility & Binary & yes, no & - & Is the detected object in the photo partially visible? \\
\bottomrule
\end{tabular}
}
\end{table*}

This comprehensive attribute set demonstrates HiBug's systematic approach through predefined visual attributes. Each attribute is designed to capture specific visual characteristics that may contribute to model failures. We provide illustrations of the results here due to the subjective nature of the process.

\subsection{GH-ESD Results on GESD Dataset}
\label{sec:fesd_detailed_results}
\subsubsection{Detailed Results on GESD Object Detection Dataset}
\label{subsec:detection_fesd_results}
\Cref{tab:fesd_detection_detailed} provides comprehensive Precision@10 results for all 21 ground truth error slices in the object detection setting, complementing the summary results presented in the main paper. GH-ESD achieves an overall average Precision@10 of 0.729, with 5 perfect matches and valid predictions for all ground truth slices.

\begin{table*}[t]
\centering
\small
\caption{\textbf{Detailed results on GESD object detection dataset.} Complete Precision@10 scores for each ground truth error slice, showing the breakdown of GH-ESD performance across different failure types (FP: False Positive, FN: False Negative).}
\label{tab:fesd_detection_detailed}
\setlength{\tabcolsep}{4pt}
\resizebox{\linewidth}{!}{
\begin{tabular}{c p{3.5cm} c p{6cm} c}
\toprule
\textbf{GT ID} & \textbf{GT Slice Name} & \textbf{GT Size} & \textbf{Best Predicted Slice} & \textbf{P@10} \\
\midrule
0 & Stroller trolley is mistakenly detected as a bicycle & 33 & folding bicycle & 0.2 \\
1 & Blurry face missed & 2265 & faces with motion blur & 0.5 \\
2 & Cars that are partially obscured missed. & 1558 & Cars that are more than 30\% obscured & 0.8 \\
3 & Nonphysical face (e.g., face reflected in a mirror) is mistakenly detected as face & 1127 & Faces on screens: a face appearing on a TV, computer monitor, or phone within the image. & 0.7 \\
4 & Paper is mistakenly detected as book & 334 & Notebooks, journals, or planners that look like a book & 0.5 \\
5 & Red traffic sign is mistakenly detected as stop sign & 2449 & red octagonal object near stop sign & \textbf{1.0} \\
6 & Person in the car missed & 249 & out of focus person & 0.9 \\
7 & Closed umbrella missed & 78 & Closed umbrella & 0.6 \\
8 & Artistic face is mistakenly detected as face & 1000 & Statues, busts, or sculptures & \textbf{1.0} \\
9 & Partial face missed & 1672 & faces partially hidden by objects & 0.8 \\
10 & Vase with no flower or plants inside missed & 136 & Vase with no flower or plants inside & 0.5 \\
11 & Baby face missed & 108 & baby faces & \textbf{1.0} \\
12 & Trailer is mistakenly detected as truck & 242 & Trailers that look like a truck & 0.7 \\
13 & Dumpster is mistakenly detected as truck & 604 & Dumpsters or large bins that look like a truck & \textbf{1.0} \\
14 & Artificial face is mistakenly detected as face & 305 & objects resembling faces & 0.4 \\
15 & Open book missed & 29 & an open book & 0.9 \\
16 & Watch missed when detecting clock & 25 & watch that looks like a clock & \textbf{1.0} \\
17 & Open suitcase missed & 12 & suitcase that will not close & 0.8 \\
18 & Bicycles seen from the front/back that being ridden/pushed by people & 335 & bicycle partially occluded by a person & 0.8 \\
19 & Cars with open doors are mistakenly detected as truck & 127 & Freight cars that look like a truck & 0.4 \\
20 & Nonvertical face in the image missed & 64 & faces with extreme head tilt & 0.8 \\
\midrule
\multicolumn{4}{l}{\textbf{Average Precision@10}} & \textbf{0.729} \\
\multicolumn{4}{l}{\textbf{Perfect Matches (P@10 = 1.0)}} & \textbf{5/21} \\
\multicolumn{4}{l}{\textbf{Valid Matches (P@10 $>$ 0)}} & \textbf{21/21} \\
\bottomrule
\end{tabular}
}
\end{table*}

The detailed results reveal GH-ESD's effectiveness across both false positive and false negative error types. The method demonstrates particularly strong performance on systematic failure patterns such as ``Red traffic sign is mistakenly detected as stop sign'' and ``Artistic face is mistakenly detected as face'' achieving perfect Precision@10 scores. Even for challenging cases like ``Nonvertical face in the image missed'', GH-ESD maintains competitive performance with Precision@10 scores of 0.8.

\subsubsection{Results on GESD Instance Segmentation Dataset}
\label{subsec:segmentation_fesd_results}

To demonstrate the generalizability of GH-ESD beyond object detection, we evaluate our method on instance segmentation tasks using the GESD dataset. 

\Cref{tab:fesd_segmentation_results} presents the detailed Precision@10 results for all 21 ground truth error slices in the instance segmentation setting. GH-ESD achieves an overall average Precision@10 of 0.805, demonstrating strong performance across diverse segmentation failure patterns. The method successfully identifies 8 perfect matches (Precision@10 = 1.0) and achieves valid matches for 20 out of 21 ground truth slices.

\begin{table*}[t]
\centering
\small
\caption{\textbf{Detailed results on GESD instance segmentation dataset.} Precision@10 scores for each ground truth error slice, showing GH-ESD's performance across diverse segmentation failure patterns.}
\label{tab:fesd_segmentation_results}
\setlength{\tabcolsep}{4pt}
\resizebox{\linewidth}{!}{
\begin{tabular}{c p{3.5cm} c p{6cm} c}
\toprule
\textbf{GT ID} & \textbf{GT Slice Name} & \textbf{GT Size} & \textbf{Best Predicted Slice} & \textbf{P@10} \\
\midrule
0 & Clustered bananas merged, boundaries unclear & 183 & banana bunch & 0.8 \\
1 & Airplane boundary expanded under occlusion & 25 & airplanes with custom paint jobs & 0.9 \\
2 & Bed over-segmented with pet & 36 & pet lying on bed & 0.9 \\
3 & Laptop keyboard merged with laptop body & 136 & laptop keyboard & \textbf{1.0} \\
4 & Paper mis-segmented as book & 334 & Notebooks, journals, or planners that look like a book & 0.7 \\
5 & Apple logo mis-segmented as apple & 34 & partially occluded apple & 0.7 \\
6 & Tie boundary unclear due to color blending & 82 & ties similar to other clothing & \textbf{1.0} \\
7 & Sliced carrot under-segmented due to shape change & 432 & bunches of carrots & 0.9 \\
8 & Adjacent toys merged with teddy bear & 81 & teddy bear with accessories & \textbf{1.0} \\
9 & Clustered oranges merged, boundaries unclear & 37 & pile of oranges & 0.7 \\
10 & Open suitcase missed (non-canonical form) & 12 & overstuffed and bulging suitcase & 0.6 \\
11 & Open book missed due to shape variation & 29 & open book & 0.6 \\
12 & Clustered apples merged, boundaries unclear & 45 & apple in fruit stand & \textbf{1.0} \\
13 & Sliced banana under-segmented due to shape change & 90 & Banana slice & 0.7 \\
14 & Closed umbrella missed (non-typical pose) & 78 & Closed umbrella & \textbf{1.0} \\
15 & Vase without flowers missed due to context prior & 136 & Vase with no flowers or plants inside & 0.6 \\
16 & Sliced orange under-segmented due to shape change & 97 & orange slice & \textbf{1.0} \\
17 & Bed over-segmented together with person & 44 & person lying on bed & \textbf{1.0} \\
18 & Surfboard on land missed (expected water context) & 36 & - & 0.0 \\
19 & Watch missed when segmenting clock & 25 & watch that looks like a clock & \textbf{1.0} \\
20 & Sliced apple under-segmented due to shape change & 82 & apple slice & 0.8 \\
\midrule
\multicolumn{4}{l}{\textbf{Average Precision@10}} & \textbf{0.805} \\
\multicolumn{4}{l}{\textbf{Perfect Matches (P@10 = 1.0)}} & \textbf{8/21} \\
\multicolumn{4}{l}{\textbf{Valid Matches (P@10 $>$ 0)}} & \textbf{20/21} \\
\bottomrule
\end{tabular}
}
\end{table*}

The results demonstrate that GH-ESD effectively transfers to instance segmentation tasks, maintaining high precision across various failure modes including occlusion patterns, object state variations, and contextual confusion scenarios. Notably, the method achieves perfect precision on challenging slices such as ``laptop keyboard merged with laptop body'' (fine-grained object parts), ``Tie boundary unclear due to color blending'' (visual similarity challenges), and shape-related patterns like ``Sliced orange under-segmented due to shape change''. The method shows robust performance across different object categories including fruits, furniture, and everyday objects, with only one slice: ``Surfboard on land missed (expected water context)'' receiving zero precision, indicating the challenge of segmenting surfboards in land-based contexts.

\subsection{Generalization to Image-level Bias Benchmark Implementation Details}
\label{subsec:image_level_details}
To get predictions for model errors, for Waterbirds and CelebA experiments we utilize pre-trained model checkpoints provided by LADDER~\cite{S-ghosh2025ladder}, which were trained with standard empirical risk minimization (ERM). 
For NICO++ experiments, we train models following the FACTS~\cite{S-yenamandra2023facts} 
, using ResNet-50 with learning rate $10^{-5}$ and weight decay sweeping over $[10^{-3}, 10^{-2}, 10^{-1}, 1.0, 2.0]$ to amplify spurious correlations. 
We report the performance metrics directly from the original papers for all compared methods. For LADDER on NICO++, where results were not available, we generated them by running the FACTS training codes~\cite{S-yenamandra2023facts} and LADDER's slice finding codes.

\subsection{Model Repair Experiment Setup Details}
\subsubsection{Object Detection Model Repair on Instance-level Patterns}
\label{subsubsec:model_details_detection}
We evaluate the actionability of discovered slices via model repair experiments on object detection tasks. We use the COCO bicycle subset (\emph{3{,}252 images, 7{,}113 bicycle instances}) and a pretrained Faster~R-CNN (ResNet-50-FPN) model. The baseline repair method is fine-tuning on all COCO images containing at least one bicycle instance. Optimization uses SGD (momentum $0.9$), weight decay $5\!\times\!10^{-4}$, batch size $16$, initial learning rate $1\!\times\!10^{-5}$, and a MultiStepLR scheduler with milestones at epochs $8$ and $11$ ($\gamma{=}0.1$) for a total of $10$ epochs.

Hypothesis matching scores of instances for the validated error slice \emph{``bicycle partially occluded by a person''} are computed by GH-ESD slice discovery method. The image-level binary attribute is defined as $\mathbf{1}\{\max_{\text{instances}} s_i \ge 0.9\}$, where $s_i$ is the instance score. Each object class defines a \emph{group} (e.g., \texttt{person}, \texttt{bicycle}). For example, a pair (\texttt{person}, True) denotes a person instance co-occurring with a likely \emph{``bicycle partially occluded by a person''} case, whereas (\texttt{bicycle}, False) represents a bicycle instance with no such occlusion in the same image. We initialize weights uniformly across all (group, attribute) pairs. Only detection-head losses (classification and box regression) are reweighted; RPN losses (objectness, RPN box) remain unweighted and are added directly to the total loss.

For the grounded repair, we adapt the online GroupDRO framework
to instance-level objectives. Each object class is treated as a group, and an image-level binary attribute is defined by thresholding the maximum instance-level hypothesis matching score at 0.9. Per-group losses are computed by aggregating proposal-level classification and regression losses within each group. The weighted detection-head loss is then

\[
\tilde{\mathcal{L}}_{\text{head}} = \frac{\sum_u q_u \mathcal{L}_u}{\sum_u q_u},
\]
where $\mathcal{L}_u$ is the mean detection loss for group $u$ and $q_u$ is its current weight. The total loss adds unweighted RPN terms:
\[
\mathcal{L}_{\text{total}} = \tilde{\mathcal{L}}_{\text{head}} + \mathcal{L}_{\text{RPN,obj}} + \mathcal{L}_{\text{RPN,box}},
\]
where $\mathcal{L}_{\text{RPN,obj}}$ and $\mathcal{L}_{\text{RPN,box}}$ are the RPN objectness and bounding box regression losses, respectively.

After each iteration, group weights are updated online as
\[
q_u \leftarrow q_u \exp(\eta \mathcal{L}_u), \quad q_u \leftarrow q_u / \sum_v q_v,
\]
with $\eta{=}10^{-5}$. The GroupDRO variant uses batch size $16$, learning rate $5\!\times\!10^{-5}$, and the same optimizer settings and scheduler as the baseline.

In summary, each iteration aggregates instance losses by group, computes a reweighted detection-head loss, adds RPN losses, and updates the group weights using the exponential rule above. Uniform weight initialization is applied at the start of fine-tuning, and all other architectural details follow the main text.

\subsubsection{Model Repair on Image Classification}
\label{subsubsec:model_details_classification}
We also conduct model repair experiments on CelebA and Waterbirds, following 
the official LADDER implementation. 
We use the pretrained ResNet weights and the author-released discovered slices provided by LADDER, and modify only the final error-mitigation stage in the official training scripts for CelebA and Waterbirds.
For LADDER, we follow its original repair procedure: each discovered slice defines a hypothesis matching score over validation samples. For every slice, we fine-tune a separate classifier using the top and bottom $n$ samples by similarity and aggregate predictions via LADDER's ensemble. 

For FACTS, Domino, and GH-ESD, the discovered slices exist only within local subsets, often containing fewer samples than required by LADDER's per-slice training. Thus, we adopt the unified Deep Feature Reweighting (DFR) framework
, which serves as the underlying procedure used in LADDER. 

For each sample, we compute a single hypothesis matching score by taking the maximum across all discovered slices, representing the likelihood of any failure pattern. 
Fine-tuning is applied only to the classifier head on the top and bottom subsets ranked by this score, producing a single repaired classifier without ensemble aggregation. All hyperparameters match the official LADDER scripts except that we randomize the random seed for each run to compute standard deviations while the original LADDER implementation uses a fixed seed.

\section{Error Slice Discovery Tool based on GH-ESD}
\label{sec:tool_implementation}

To demonstrate the practical utility of GH-ESD, we developed a production-ready tool for error slice discovery in computer vision systems. While the GH-ESD described in our paper is an automated pipeline given a task context for evaluation purposes, the deployed tool is designed for an interactive workflow. It provides a user interface that allows human expertise to be incorporated into each stage of the discovery process.
The following screenshots illustrate the key components of our implementation across different steps on the Waterbirds dataset.
\subsection{Interactive Chat Interface for Hypothesis Construction}

\begin{figure*}[t]
    \centering
    \includegraphics[width=\linewidth]{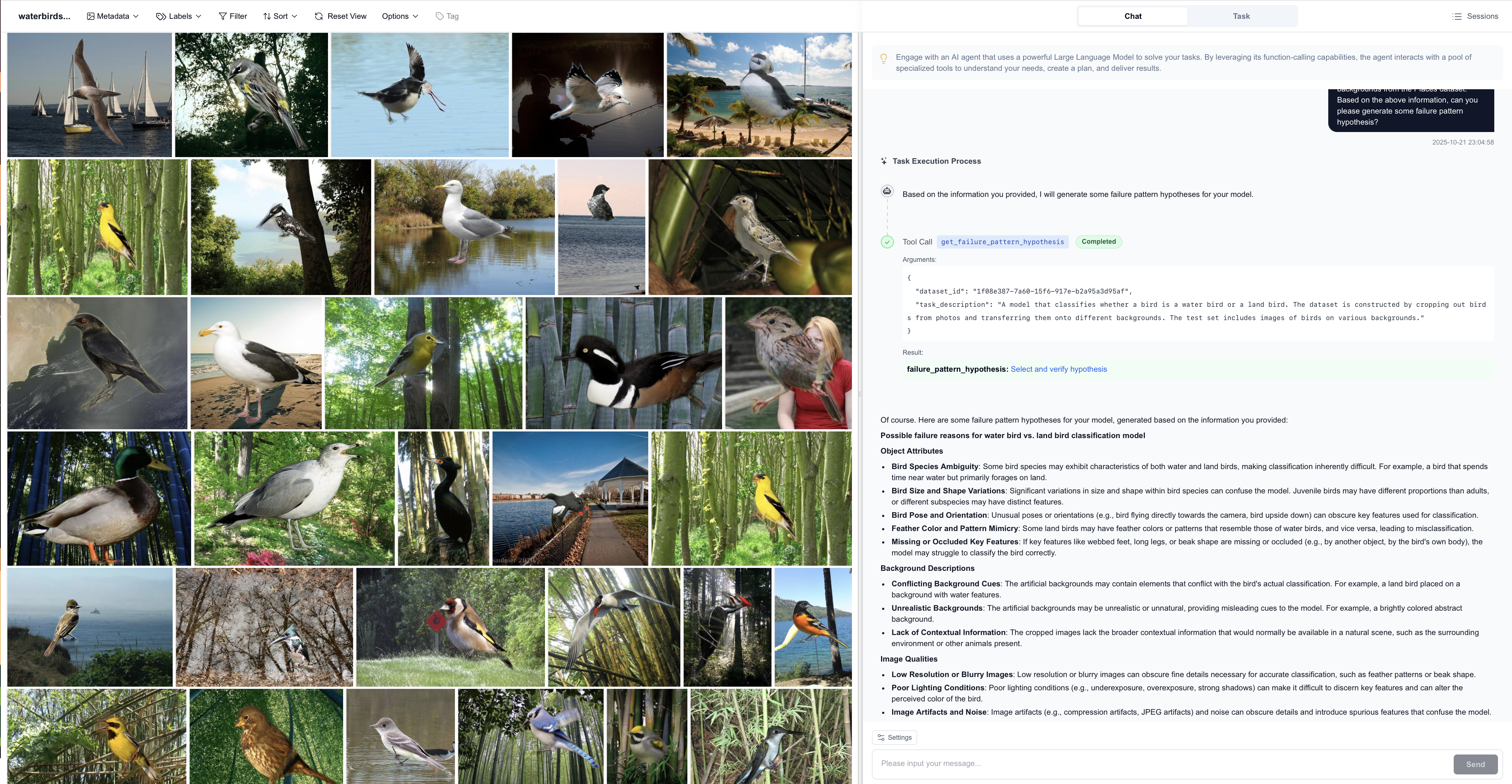}
    \caption{\textbf{Main tool interface with interactive chat functionality.} The left panel displays the data gallery for dataset visualization, while the right panel provides a conversational interface for users to interact with the LLM for hypothesis generation and error slice discovery tasks.}
    \label{fig:chatbox_hypothesis}
\end{figure*}

\Cref{fig:chatbox_hypothesis} shows the main interface of the tool, featuring a dual-panel design with a data gallery on the left for dataset visualization and an interactive chatbox on the right for conversational hypothesis construction and data retrieval.

\subsection{Hypothesis Selection and Submission}

\begin{figure*}[t]
    \centering
    \includegraphics[width=\linewidth]{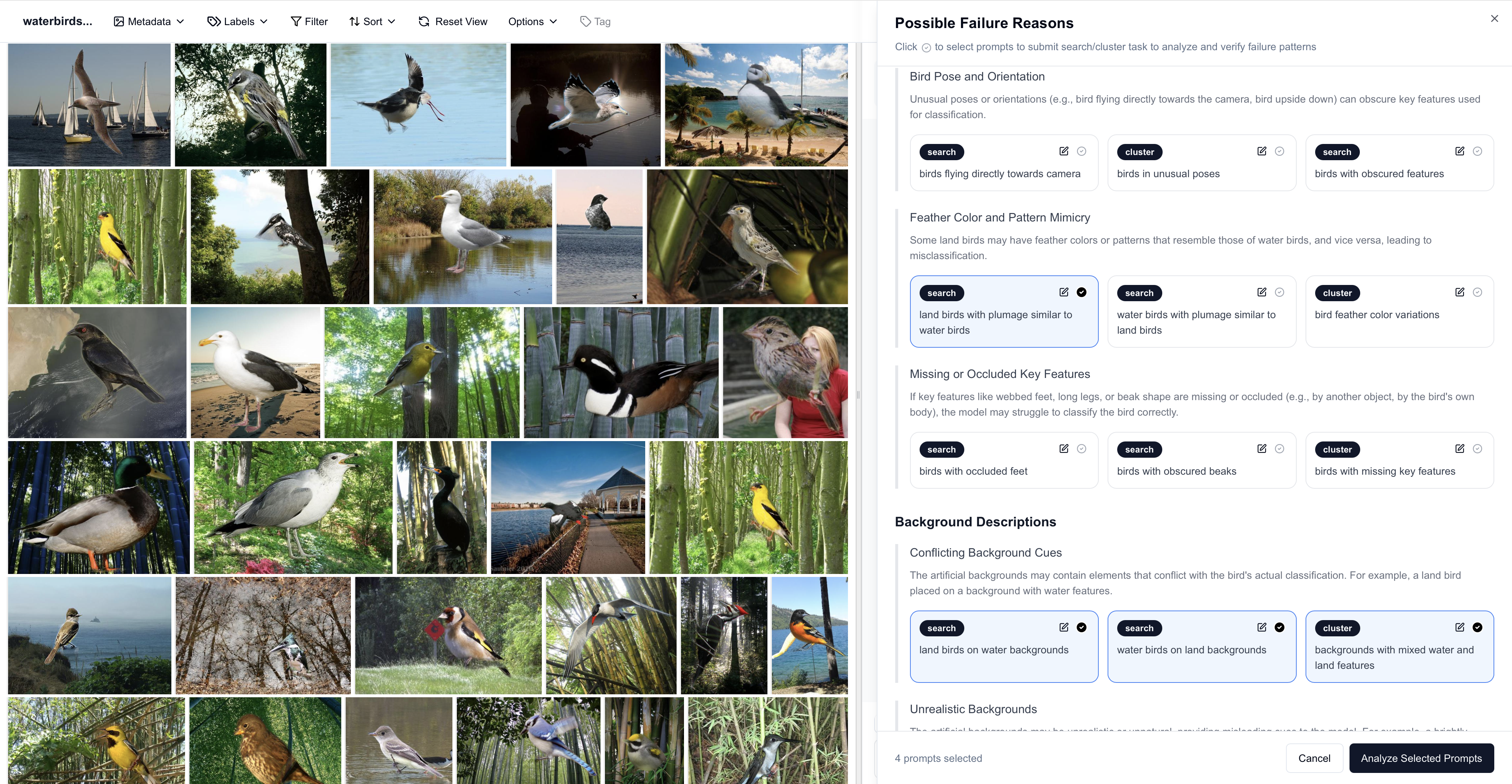}
    \caption{\textbf{Hypothesis selection and submission interface.} Users can review automatically generated failure pattern hypotheses, assess their plausibility, and submit selected candidates for validation.}
    \label{fig:submit_hypothesis}
\end{figure*}

\Cref{fig:submit_hypothesis} demonstrates the hypothesis selection interface, where users can review GH-ESD constructed failure pattern hypotheses and select the most promising candidates. The interface allows users to submit selected hypotheses for instance-level matching and statistical verification.

\subsection{Hypothesis Verification Task Management}

\begin{figure*}[t]
    \centering
    \includegraphics[width=\linewidth]{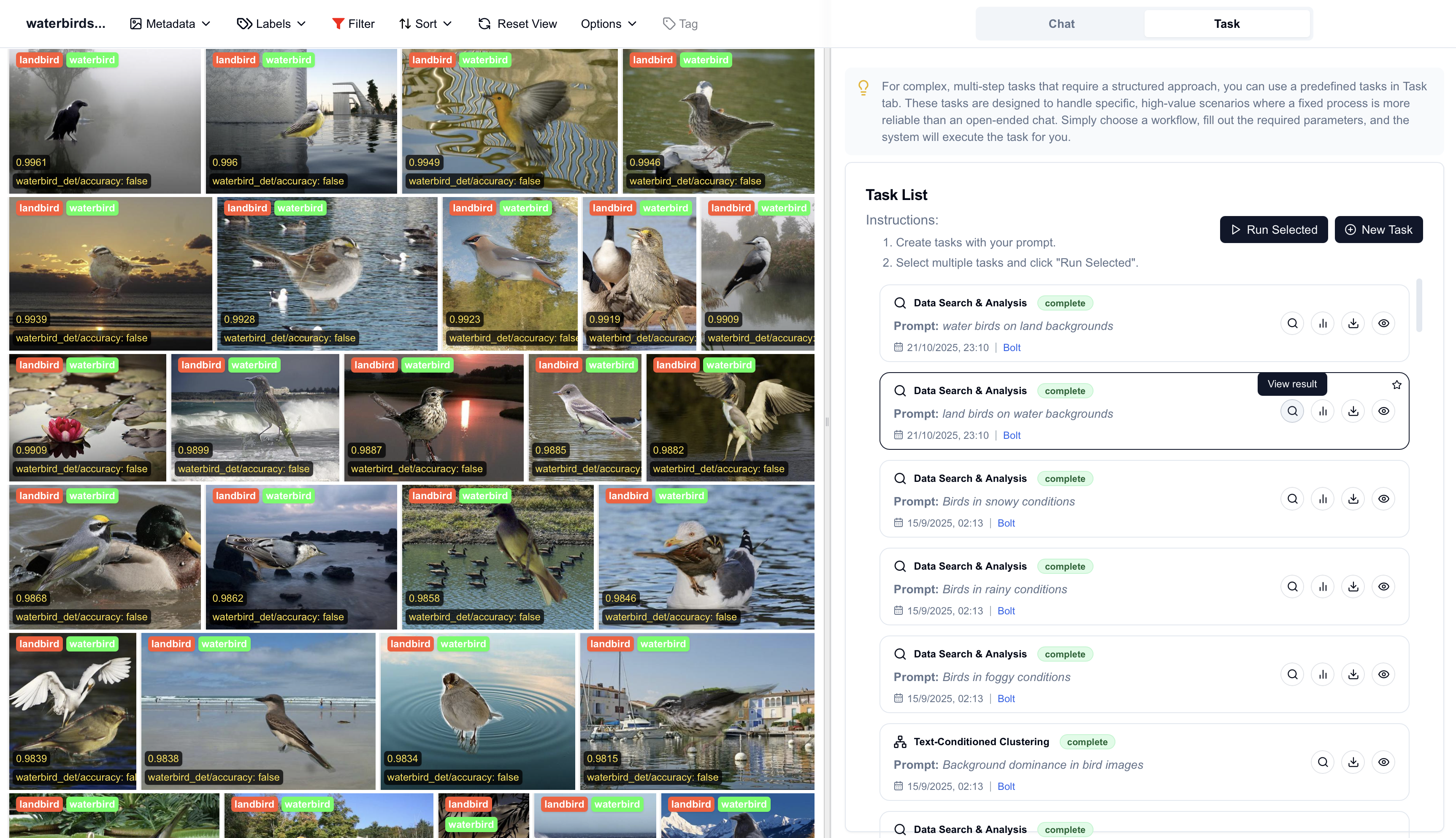}
    \caption{\textbf{Hypothesis verification task management interface.} The interface displays all submitted validation tasks with their current status, allowing users to track progress and access completed results including hypothesis matching and statistical verification results.}
    \label{fig:hypothesis_list}
\end{figure*}

\Cref{fig:hypothesis_list} presents the task management interface, which displays all submitted hypotheses along with their validation status. Each hypothesis-validation task runs asynchronously, allowing users to monitor progress and access results upon completion. These tasks are executed in batches to improve performance through prefilling and batched request optimization in the vLLM framework.

\subsection{Error Slice Hypothesis Verification}

\begin{figure*}[t]
    \centering
    \includegraphics[width=\linewidth]{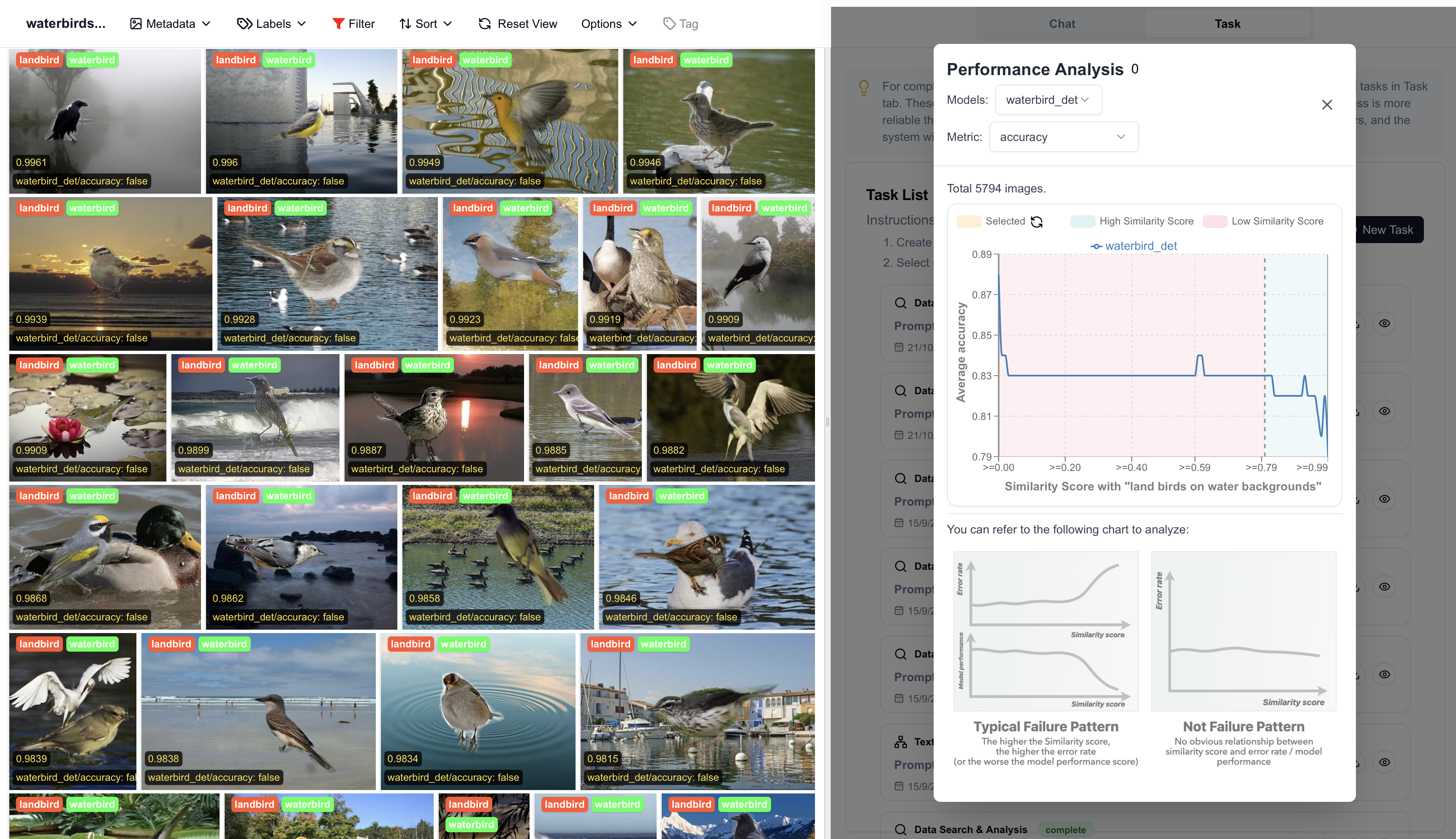}
    \caption{\textbf{Statistical hypothesis verification results interface.} Example showing the statistical hypothesis verification through monotonic trend analysis between hypothesis matching scores and model performance, demonstrating robust evidence for discovered error slices. Note that the Y-axis in this example is \textit{model accuracy}, not \textit{model error rate}; therefore, a downward trend (rather than an upward trend) of the curve suggests it is a failure pattern.}
    \label{fig:correlation_analysis}
\end{figure*}

\Cref{fig:correlation_analysis} shows an example of the statistical analysis results for a validated hypothesis. The interface presents the statistical relationship between hypothesis matching scores and performance metric (e.g., accuracy). Unlike experiments in the paper, which report error rates as the performance metric, the tool supports analysis using either accuracy or error rate, and the analysis adapts automatically to the selected metric. This provides robust evidence supporting the validity of the discovered error slices across multiple metrics.

%

\end{document}